\documentclass[11pt]{article}
\usepackage[preprint]{acl}

\usepackage{latexsym}
\usepackage{microtype}
\usepackage{graphicx}

\usepackage[english]{babel}
\babelfont{rm}[Extension=.otf,
  UprightFont=*-regular, BoldFont=*-bold,
  ItalicFont=*-italic, BoldItalicFont=*-bolditalic]{texgyretermes}

\usepackage{xeCJK}
\xeCJKDeclareCharClass{Default}{"2014, "2018, "2019, "201C, "201D}
\setCJKsansfont{HaranoAjiGothic-Regular.otf}[BoldFont=HaranoAjiGothic-Bold.otf]

\newCJKfontfamily\foreignjfont{FandolSong-Regular.otf}
\newfontfamily\foreignfont{DejaVuSans.ttf}
\providecommand{\foreign}[1]{{\upshape\foreignjfont\foreignfont\textcolor{red}{#1}}}

\usepackage{comment}
\usepackage{xcolor}
\definecolor{airportcolor}{RGB}{40,143,145}
\definecolor{namecolor}{RGB}{75,42,140}

\usepackage[normalem]{ulem}
\newcommand{\reduwave}[1]{{\color{red}\uwave{{\color{black}#1}}}}

\usepackage{multirow}
\usepackage{multicol}
\usepackage{booktabs}
\usepackage{amsmath}
\usepackage{longtable}
\usepackage{makecell}
\usepackage{enumitem}
\usepackage{nicematrix}
\usepackage{tcolorbox}
\usepackage{afterpage}
\usepackage{amssymb}
\usepackage{tabularx}
\usepackage{newunicodechar}
\newunicodechar{✓}{\ensuremath{\checkmark}}
\newcolumntype{L}{>{\raggedright\arraybackslash}X}

\newcommand{\sd}[1]{{\scriptsize $\pm$#1}}

\newcommand{\agent}[1]{\textcolor{black}{#1}}
\newcommand{\user}[1]{\textcolor{black}{#1}}

\newcommand{\agenttoolcall}[1]{\textcolor{black}{#1}}
\newcommand{\toolresp}[1]{\textcolor{black}{#1}}
\newlength{\dwd}
\newcommand{\dup}[1]{\,\makebox[\dwd][l]{\scriptsize\textcolor{red}{(+#1)}}}
\newcommand{\ddn}[1]{\,\makebox[\dwd][l]{\scriptsize\textcolor{blue}{($-$#1)}}}
\newcommand{\deq}[1]{\,\makebox[\dwd][l]{\scriptsize\textcolor{gray}{($\pm$#1)}}}
\setlist[enumerate]{
  topsep=1pt,     
  partopsep=0pt,  
  itemsep=0.5pt,  
  parsep=0pt      
}

\newcommand\blfootnote[1]{
  \begingroup
  \renewcommand\thefootnote{}\footnote{#1}
  \addtocounter{footnote}{-1}
  \endgroup
}

\title{J-tau: A Fully Localized Japanese Benchmark \\for Tool-Agent-User Interaction}
\author{
\textbf{Chihiro Yano\textsuperscript{1,2}}\thanks{Work done during an internship at SB Intuitions Corp.},
 \textbf{Jun Hirako\textsuperscript{1}},
 \textbf{Hiroaki Yamagiwa\textsuperscript{1}},
 \textbf{Ryota Hirobuchi\textsuperscript{1}}
\\
 \textsuperscript{1}SB Intuitions Corp.,
 \textsuperscript{2}Nagoya University
\\
\texttt{\{chihiro.yano, jun.hirako\}@sbintuitions.co.jp},
\\
\texttt{\{hiroaki.yamagiwa, ryota.hirobuchi\}@sbintuitions.co.jp}
\\
}

\begin{document}
\maketitle
\begin{abstract}
We present J-tau, the first publicly available Japanese benchmark for evaluating tool-using LLM agents in interactive customer-support scenarios. 
While tau-bench is widely used to evaluate LLM agents, it is English-centric and cannot evaluate agents in other languages. 
Existing multilingual extensions offer no public Japanese version and rely primarily on machine translation, leaving locale-specific entities such as addresses and currencies inconsistent with the target region. 
To enable reliable evaluation in Japanese, we carefully localize tau-bench through machine translation, entity adaptation, and human review. During this process, we also uncover and fix two flaws in the original tau-bench: invalid tasks and a score aggregation error that inflates certain models’ results. 
Experiments show that J-tau yields fewer evaluation errors than a naively machine-translated counterpart and that our fixes resolve both flaws.\blfootnote{We release J-tau at \url{https://github.com/sbintuitions/j-tau-bench}.}
\end{abstract}

\section{Introduction}
\begin{figure}[t]
    \centering
    \includegraphics[width=0.95\linewidth]{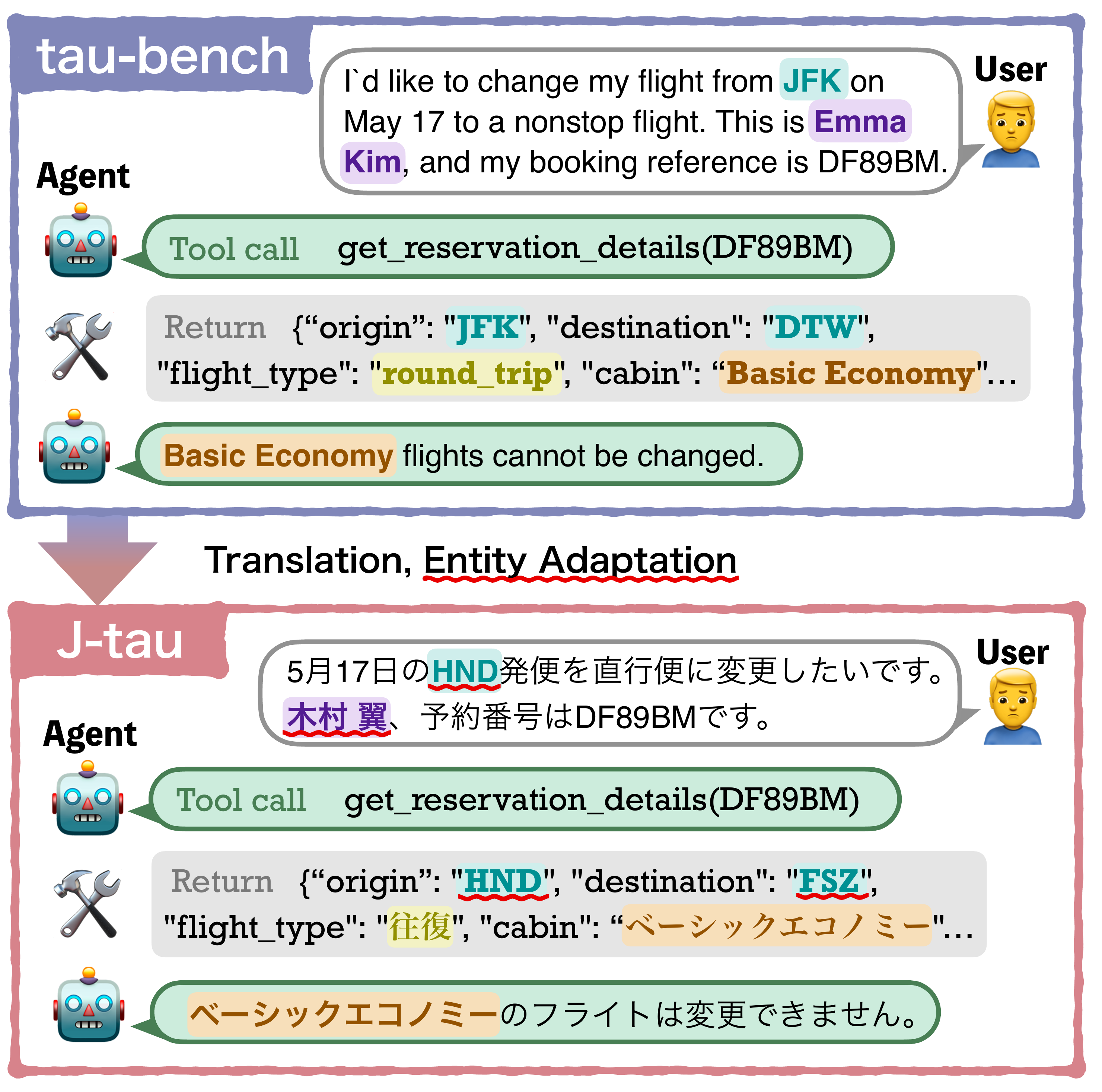}
    \caption{
Example of Japanese localization in J-tau.
We translate the original tau-bench and replace locale-specific entities (marked with \reduwave{wavy underlines}) with realistic Japanese counterparts.
Examples include airport codes (\textcolor{airportcolor}{JFK} $\rightarrow$ \textcolor{airportcolor}{HND}, \textcolor{airportcolor}{DTW} $\rightarrow$ \textcolor{airportcolor}{FSZ}) and personal names (\textcolor{namecolor}{Emma Kim} $\rightarrow$ \textcolor{namecolor}{木村 翼}, where Kimura (木村) is the family name and Tsubasa (翼) is the given name).
}
    \label{fig:teaser}
\end{figure}
Large language model (LLM) agents that autonomously use tools to solve complex tasks have attracted growing attention~\cite{DBLP:journals/fcsc/WangMFZYZCTCLZWW24}.
While many benchmarks have been proposed to evaluate such agents, most of them focus on English. 
Because agent performance can vary across languages~\cite{maps}, English evaluations may not reflect utility in other languages.

To address this gap, multilingual agent benchmarks have been proposed, including ones built from multilingual corpora and machine translations of English benchmarks~\cite{massive_agents,seatau}. 
However, machine translation alone cannot localize entities such as personal names, which can lead to unnatural conversations or even invalid tasks~\cite{gaia_lilt}.

\begin{figure*}[t]
    \centering
    \includegraphics[width=0.91\linewidth]{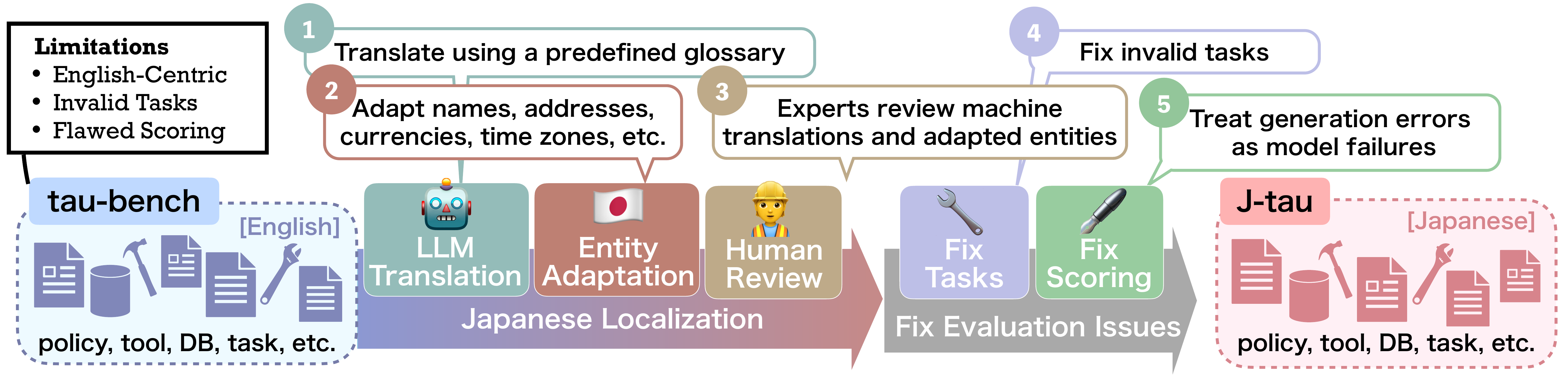}
\caption{
Overview of the J-tau pipeline, which localizes tau-bench into Japanese and fixes its evaluation issues.
}
    \label{fig:Construction}
\end{figure*}

To overcome these limitations, we present and release J-tau, a Japanese extension of the widely used English benchmark $\tau$-bench~(tau-bench)~\cite{taubench,tau2bench,tau-fix}. 
J-tau is the first publicly available Japanese benchmark that evaluates tool-using LLM agents in realistic, interactive scenarios. 
It features high-quality localization (Fig.~\ref{fig:teaser}) and corrected tasks and evaluation.
Our main contributions are as follows:

\begin{itemize}[leftmargin=*, itemsep=2pt, parsep=0pt, topsep=0pt]
\item We construct and release J-tau, a Japanese benchmark that localizes three domains of tau-bench.
\item We identify two flaws in tau-bench and fix them in J-tau: invalid tasks (about 10\% in airline and retail) and score-inflating aggregation (up to 64.6 points for a single model).
\item We show that an existing machine-translation-based pipeline often causes evaluation failures that stem from the framework rather than agents.
\end{itemize}

\section{Background: tau-bench}

This section describes tau-bench and its limitations, along with prior work addressing them.\footnote{See
Appendix~\ref{app:related_work} for additional related work.}

\subsection{Overview}
tau-bench is a benchmark for evaluating LLM agents on customer support tasks across multiple domains, including airline (booking flights), retail (modifying orders), and telecom (troubleshooting mobile phones).
In each task, an LLM-simulated user follows a user instruction describing its goal, and the agent must resolve the request conversationally, calling tools while complying with a domain-specific policy.
In telecom, as in real-world technical support, the user has a separate set of tools, so the agent must also guide the user in using them.

\subsection{Limitations}
\label{sec:taubench_limitations}
tau-bench has three main limitations.
First, it is English-centric. 
While prior work extends it to five other languages\footnote{Mandarin, Vietnamese, Thai, Indonesian, and Filipino.}~\cite{seatau}, this extension relies primarily on machine translation, which degrades data quality and leads to unnatural conversations.\footnote{For example, users converse in a Southeast Asian language, yet addresses and currencies remain those of the U.S.}
Second, although tau-bench has been revised since its release~\cite{tau-fix}, it still contains invalid tasks that are unsolvable or have non-deterministic ground truths.
Third, its score aggregation is flawed and overestimates the performance of some models.

\section{Japanese Agent Benchmark: J-tau}
This section describes the construction pipeline and details of J-tau, a Japanese agent benchmark that addresses the limitations of tau-bench.\footnote{Except for the language, evaluation fixes, and user simulator, we follow tau-bench's settings (e.g., prompts).}

\subsection{Construction Pipeline}\label{sec:construction-pipeline}
Our pipeline consists of five steps (see Fig.~\ref{fig:Construction}): three for Japanese localization (translation, entity adaptation, and human review) and two for fixing evaluation issues (task fixes and scoring fixes).

\paragraph{Step 1: LLM Translation.}
We translate the policies, tool definitions, databases, and user instructions from English into Japanese using Gemini 3.1 Pro~\cite{gemini}.
To ensure consistent translation of technical terms, we include a predefined domain-specific glossary in the prompt.
Appendix~\ref{app:term_dic} shows the domain-specific glossaries.

\paragraph{Step 2: Entity Adaptation.}
Direct translation retains locale-specific entities from English-speaking regions, such as personal names, addresses, and phone numbers, making the simulated conversations unnatural for Japanese speakers.
We therefore replace them with fictitious Japanese counterparts.
We generate them with \texttt{Faker}\footnote{\url{https://github.com/joke2k/faker}} for the retail and airline domains, which contain many users, and create them manually for the telecom domain.
We also convert monetary amounts into Japanese yen (1 USD = 150 JPY) and the time zone from EST to JST (see Appendix~\ref{app:detail_entity} for more details).

\paragraph{Step 3: Human Review.}
We then assess and correct the translation quality of the resulting data.
For each domain, three experts review the policies, and for the airline and retail domains, experts also review the user instructions (see Appendix~\ref{app:detail_annotation}).

\paragraph{Step 4: Fix Tasks.}
We verify each task through manual inspection and test runs, and identify tasks that cannot be solved.\footnote{We verify only the airline and retail domains, since telecom tasks are programmatically generated and always solvable.}
We find two types of problematic tasks: tasks that become unsolvable through localization, and invalid tasks already present in the original tau-bench, which are unsolvable or have non-deterministic ground truths.\footnote{For example, when the user instruction lacks the information needed to choose among several payment cards, the agent must pick one, and whether it matches the card in the reference solution is a matter of chance rather than model capability.}
We fix these tasks in J-tau, for example by adding missing information to the user instructions (see Appendix~\ref{app:fixed_task}).

\paragraph{Step 5: Fix Scoring.}
tau-bench treats both empty responses and tool calls with malformed JSON as infrastructure errors. It retries them up to three times and excludes trials with persistent errors from aggregation. However, agent errors reflect the agent's limitations, so this procedure inflates the scores of models prone to such errors.\footnote{For example, although Qwen3.5-4B ended 75/150, 56/342, and 38/342 trials with empty responses in airline, retail, and telecom, respectively, it scored 93.3, 85.3, and 98.7.} In J-tau, we instead attribute these errors to the model that produced them. Trials are counted as failures for agent errors and discarded for user simulator errors.

\subsection{Benchmark Details}
\label{sec:j-tau-detail}
Constructed through this pipeline, J-tau is a high-quality Japanese agent benchmark covering the airline, retail, and telecom domains.\footnote{Appendix~\ref{app:stats_jtau} reports the number of tasks in each domain.}

In interactive agent evaluation, the choice of user simulator can substantially affect agent scores~\cite{lost_in_sim}.
We therefore compare several recent LLMs in terms of Japanese naturalness and instruction-following stability, and adopt GLM-5.2-FP8~\cite{glm5}, which distorted evaluation results the least, as the default user simulator (see Appendix~\ref{app:user_select}).

\begin{table}[t]
    \centering
    \small
    \begin{tabular}{@{\hspace{0.2em}}l@{\hspace{0.8em}}r@{\hspace{0.8em}}r@{\hspace{0.8em}}r@{\hspace{0.2em}}}
        \toprule
         Model & airline & retail & telecom \\
        \midrule
        \multicolumn{2}{l}{ Proprietary Models}  \\
        \midrule
         GPT-5.6 Sol            & 84.7\sd{2.3} & 83.0\sd{4.0} & 90.9\sd{1.8} \\
        Gemini 3.1 Pro         & 84.0\sd{4.0} & 78.9\sd{3.2} & 79.5\sd{4.5} \\
        Claude Opus 4.8        & 81.3\sd{3.1} & 77.5\sd{2.8} & 84.5\sd{2.2} \\
        Claude Fable 5         & 90.7\sd{1.2} & 84.5\sd{4.1} & 99.4\sd{0.5} \\
        \midrule
        \multicolumn{2}{l}{ Open-Weight Models}  \\
        \midrule
        Kimi-K3 & 70.0\sd{2.0} & 49.1\sd{7.0} & 67.0\sd{3.5} \\
        GLM-5.2-FP8 & 89.3\sd{4.6} & 81.6\sd{1.8} & 98.2\sd{1.5} \\
        gpt-oss-120b & 73.3\sd{4.2} & 75.7\sd{0.5} & 64.6\sd{1.0} \\
        gpt-oss-20b & 50.7\sd{3.1} & 51.2\sd{3.3} & 59.1\sd{3.3} \\
        Qwen3.5-397B-A17B-FP8 & 83.3\sd{6.4} & 85.1\sd{2.3} & 96.2\sd{0.5} \\
        Qwen3.5-27B & 84.7\sd{4.2} & 75.1\sd{1.3} & 89.8\sd{8.7} \\
        Qwen3.5-9B & 49.3\sd{2.3} & 35.7\sd{7.3} & 50.0\sd{9.2} \\
        Qwen3.5-4B & 24.0\sd{2.0} & 6.7\sd{2.2}  & 33.6\sd{9.9} \\
        gemma-4-31b-it & 51.3\sd{6.4} & 72.8\sd{4.9} & 44.4\sd{9.3} \\
        gemma-4-12b-it & 41.3\sd{3.1} & 34.5\sd{3.5} & 23.1\sd{5.1} \\
        gemma-4-E4B-it & 38.7\sd{5.0} & 39.2\sd{5.3} & 20.2\sd{5.8} \\
        gemma-4-E2B-it & 20.7\sd{4.2} & 26.3\sd{0.9} & 13.2\sd{4.0} \\
        llm-jp-4.1-33b-thinking & 64.7\sd{3.1} & 68.4\sd{2.6} & 72.5\sd{5.1} \\
        llm-jp-4.1-8b-thinking & 56.0\sd{3.5} & 63.7\sd{3.7} & 52.0\sd{5.6} \\
        \bottomrule
    \end{tabular}
    \caption{Results on J-tau. We report the mean and standard deviation over three trials.}
    \label{tab:main_result}
\end{table}

\section{Evaluating LLM Agents on J-tau}
In this section, we evaluate and analyze a range of proprietary and open-weight models on J-tau.

\paragraph{Setup.}
We run three trials per task and report the mean and standard deviation.\footnote{For the retail domain, we use \texttt{gpt-oss-120b}~\cite{gpt-oss} as the LLM judge for conversation-based evaluation.}
Appendix~\ref{app:models} provides details of the evaluated models.

\begin{table*}[t!]
\centering
\footnotesize
\begin{tabular}{@{\hspace{0.1em}}l@{\hspace{0.5em}}r@{\hspace{0.7em}}r@{\hspace{0.95em}}r@{\hspace{0.7em}}r@{\hspace{0.95em}}r@{\hspace{0.7em}}r@{\hspace{0.95em}}r@{\hspace{0.7em}}r@{\hspace{0.95em}}r@{\hspace{0.7em}}r@{\hspace{0.1em}}}
\toprule
\multirow{4}{*}{Model} & \multicolumn{4}{c}{airline} & \multicolumn{4}{c}{retail} & \multicolumn{2}{c}{telecom} \\
\cmidrule(lr){2-5} \cmidrule(lr){6-9} \cmidrule(lr){10-11}
 & \multicolumn{2}{c}{Repaired$_{\,(6)}$} & \multicolumn{2}{c}{Unmodified$_{\,(41)}$} & \multicolumn{2}{c}{Repaired$_{\,(12)}$} & \multicolumn{2}{c}{Unmodified$_{\,(80)}$} & \multicolumn{2}{c}{Unmodified$_{\,(114)}$} \\
\cmidrule(lr){2-3} \cmidrule(lr){4-5} \cmidrule(lr){6-7} \cmidrule(lr){8-9} \cmidrule(lr){10-11}
 & J-tau & $\text{tau}^*$ & J-tau & $\text{tau}^*$ & J-tau & $\text{tau}^*$ & J-tau & $\text{tau}^*$ & J-tau & $\text{tau}^*$ \\
\midrule
GPT-5.6 Sol & 77.8\dup{50.0} & 27.8 & 85.4\ddn{9.8} & 95.1 & 61.1\dup{2.8} & 58.3 & 85.4\dup{2.9} & 82.5 & 90.9\dup{5.8} & 85.1 \\
Kimi-K3 & 50.0\dup{5.6} & 44.4 & 73.2\ddn{13.8} & 87.0 & 44.4\ddn{11.1} & 55.6 & 47.5\ddn{36.7} & 84.2 & 67.0\ddn{21.9} & 88.9 \\
GLM-5.2-FP8 & 77.8\dup{44.4} & 33.3 & 91.1\ddn{3.3} & 94.3 & 77.8\dup{33.3} & 44.4 & 80.4\ddn{2.5} & 82.9 & 98.2\deq{0.0} & 98.2 \\
gpt-oss-120b & 44.4\dup{22.2} & 22.2 & 77.2\ddn{4.9} & 82.1 & 55.6\ddn{13.9} & 69.4 & 76.7\ddn{9.6} & 86.3 & 64.6\ddn{28.1} & 92.7 \\
gpt-oss-20b & 11.1\ddn{22.2} & 33.3 & 55.3\ddn{13.0} & 68.3 & 36.1\ddn{5.6} & 41.7 & 45.4\ddn{20.4} & 65.8 & 59.1\ddn{17.5} & 76.6 \\
Qwen3.5-397B-A17B-FP8 & 72.2\dup{27.8} & 44.4 & 83.7\ddn{9.8} & 93.5 & 88.9\dup{19.4} & 69.4 & 82.9\ddn{7.1} & 90.0 & 96.2\ddn{2.3} & 98.5 \\
Qwen3.5-27B & 66.7\dup{16.7} & 50.0 & 86.2\deq{0.0} & 86.2 & 75.0\dup{8.3} & 66.7 & 71.3\ddn{14.6} & 85.8 & 89.8\ddn{7.6} & 97.4 \\
Qwen3.5-9B & 22.2\dup{5.6} & 16.7 & 52.8\ddn{30.1} & 82.9 & 16.7\ddn{38.9} & 55.6 & 35.4\ddn{36.7} & 72.1 & 50.0\ddn{30.1} & 80.1 \\
Qwen3.5-4B & 0.0\ddn{16.7} & 16.7 & 27.6\ddn{2.4} & 30.1 & 0.0\ddn{27.8} & 27.8 & 5.0\ddn{47.5} & 52.5 & 33.6\ddn{18.4} & 52.0 \\
llm-jp-4.1-33b-thinking & 33.3\ddn{11.1} & 44.4 & 71.5\dup{0.8} & 70.7 & 55.6\ddn{5.6} & 61.1 & 67.9\ddn{7.9} & 75.8 & 72.5\ddn{8.2} & 80.7 \\
llm-jp-4.1-8b-thinking & 11.1\ddn{11.1} & 22.2 & 61.8\ddn{2.4} & 64.2 & 61.1\dup{2.8} & 58.3 & 61.3\ddn{5.0} & 66.3 & 52.0\ddn{23.4} & 75.4 \\
\bottomrule
\end{tabular}%

\caption{
Effect of task fixes on Repaired and Unmodified tasks (mean over three trials).
Differences of J-tau from $\text{tau}^*$ are shown in \textcolor{red}{red} (higher) and \textcolor{blue}{blue} (lower).
Subscripts next to group names denote task counts.
}
\label{tab:lang_effect}
\end{table*}

\begin{table}[t]
\centering
\footnotesize
\setlength{\tabcolsep}{4.5pt}
\begin{tabular}{@{\hspace{0.1em}}l@{\hspace{0.35em}}r@{\hspace{0.5em}}r@{\hspace{0.5em}}r@{\hspace{0.5em}}r@{\hspace{0.5em}}r@{\hspace{0.5em}}r@{\hspace{0.1em}}}
\toprule
\multirow{2}{*}{Model} & \multicolumn{2}{c}{airline} & \multicolumn{2}{c}{retail} & \multicolumn{2}{c}{telecom} \\
\cmidrule(lr){2-3} \cmidrule(lr){4-5} \cmidrule(lr){6-7}
 & $\text{tau}^*$ & tau & $\text{tau}^*$ & tau & $\text{tau}^*$ & tau \\
\midrule
GPT-5.6 Sol & 87.3 & 85.3 & 81.3 & 84.5 & 85.1 & 82.7 \\
Kimi-K3 & 80.0 & 72.7 & 82.5 & 83.0 & 88.9 & 91.2 \\
GLM-5.2-FP8 & 87.3 & 89.3 & 80.7 & 84.2 & 98.2 & 98.8 \\
gpt-oss-120b & 73.3 & 76.7 & 85.7 & 81.0 & 92.7 & 92.4 \\
gpt-oss-20b & 62.7 & 68.0 & 64.3 & 75.7 & 76.6 & 83.0 \\
Qwen3.5-397B-A17B-FP8 & 86.7 & 86.7 & 88.9 & 88.3 & 98.5 & 99.7 \\
Qwen3.5-27B & 81.3 & 84.7 & 85.4 & 86.5 & 97.4 & 99.1 \\
Qwen3.5-9B & 74.0 & 77.3 & 72.2 & 86.7 & 80.1 & 96.5 \\
Qwen3.5-4B & 28.7 & 93.3 & 50.9 & 85.3 & 52.0 & 98.7 \\
llm-jp-4.1-33b-thinking & 66.0 & 64.7 & 74.6 & 78.4 & 80.7 & 83.6 \\
llm-jp-4.1-8b-thinking & 57.3 & 60.0 & 68.7 & 68.7 & 75.4 & 73.7 \\
\bottomrule
\end{tabular}
\caption{Effect of the scoring fix, comparing tau and $\text{tau}^*$ for each domain (mean over three trials).}
\label{tab:framework_effect}
\end{table}

\begin{table}[t]
  \centering
  \small
  \setlength{\tabcolsep}{2.5pt}
    \setlength{\aboverulesep}{0.6pt}  
        \setlength{\belowrulesep}{0.8pt}  

\begin{NiceTabular}{@{\hspace{0.1em}}l@{\hspace{0.4em}}l@{\hspace{0.4em}}l@{\hspace{0.7em}}r@{\hspace{0.5em}}r@{\hspace{0.5em}}r@{\hspace{0.5em}}r@{\hspace{0.1em}}}
\toprule
\multirow{2}{*}{Model} & \multirow{2}{*}{Domain} & \multirow{2}{*}{Bench} & \multirow{2}{*}{Pass}  & \multicolumn{3}{c}{Fail} \\
\cmidrule(lr){5-7}
& & &  & \textit{agent} & \textit{user} & \textit{eval} \\
\midrule
    \multirow{6}{*}{\makecell[c]{GLM-5.2\\-FP8}} & \multirow{2}{*}{airline} & J-tau & 134 & 7 & 5 & 4 \\
    & & $\text{SEATau}_{\text{ja}}$ & 136 & 6 & 1 & 7 \\ \cmidrule{2-7}
    &\multirow{2}{*}{retail} & J-tau & 279 & 23 & 36 & 4 \\
    & & $\text{SEATau}_{\text{ja}}$ & 256 & 18 & 14 & 54 \\ \cmidrule{2-7}
    &\multirow{2}{*}{telecom} & J-tau & 336 & 4 & 2 & 0 \\
    & & $\text{SEATau}_{\text{ja}}$ & 331 & 10 & 0 & 1 \\ 
    \midrule
    \multirow{6}{*}{\makecell[c]{Qwen3.5\\-397B-A17B\\-FP8}} & \multirow{2}{*}{airline} & J-tau & 125 & 20 & 3 & 2 \\
     & & $\text{SEATau}_{\text{ja}}$ & 121 & 20 & 5 & 4 \\ \cmidrule{2-7}
     &\multirow{2}{*}{retail} & J-tau & 291 & 29 & 16 & 6 \\
      && $\text{SEATau}_{\text{ja}}$ & 236 & 36 & 14 & 56 \\ \cmidrule{2-7}
     &\multirow{2}{*}{telecom} & J-tau & 329 & 11 & 0 & 2 \\
      && $\text{SEATau}_{\text{ja}}$ & 329 & 12 & 1 & 0 \\
    \bottomrule
    
  \end{NiceTabular}
\caption{
Successful (Pass) and failed (Fail) trial counts on J-tau and $\text{SEATau}_{\text{ja}}$.
GPT-5.6 Sol~\cite{gpt56sol} classifies each failure's cause as \textit{agent}, \textit{user}, or \textit{eval}.
}
\label{tab:seatau}
\end{table}

\paragraph{Quantitative Results.}
Table~\ref{tab:main_result} shows the results for each domain of J-tau.
Among proprietary models, Claude Fable 5~\cite{claudefable} achieves the highest overall score.
Among open-weight models, GLM-5.2-FP8 and Qwen3.5-397B-A17B-FP8~\cite{qwen3.5} perform best.
Within the Qwen3.5 series, performance generally improves with model size.

\paragraph{Case Study.}
Appendix~\ref{app:case_study} presents example conversations from GLM-5.2-FP8, the open-weight model with the highest J-tau score, and failure cases from gemma-4-E2B-it~\cite{gemma4} and Qwen3.5-4B, the two lowest-scoring ones.
For example, Qwen3.5-4B sometimes switches from Japanese to another language partway through the conversation and fails to recognize the family-name-first order of Japanese names, such as Kimura Tsubasa (\textcolor{namecolor}{木村 翼}) in Fig.~\ref{fig:teaser}.

\section{Quality Analysis of J-tau}

\subsection{Effect of Task and Scoring Fixes}
\label{sec:tau*}
We conduct an ablation study to assess the effectiveness of the task and scoring fixes in our pipeline in Section~\ref{sec:construction-pipeline}.
To this end, we use $\text{tau-bench}^*$~($\text{tau}^*$), a variant of tau-bench~(tau) to which only the scoring fix in Step 5 is applied.

\paragraph{Fix Tasks.}
To examine the effect of the task fixes in Step 4, we split tasks into Repaired and Unmodified groups.
Table~\ref{tab:lang_effect} compares the J-tau and $\text{tau}^*$ scores of each group for each domain.
On the Repaired tasks, high-performing models score substantially higher on J-tau than on $\text{tau}^*$.
This is because the invalid tasks in $\text{tau}^*$ cap the achievable accuracy regardless of model capability, and our fixes remove this cap.
On the Unmodified tasks, which allow a direct comparison between languages, J-tau scores tend to be lower than $\text{tau}^*$ scores.
This is consistent with \citet{maps}, who show that agent performance varies across languages.

\paragraph{Fix Scoring.}
To examine the effect of the scoring fix in Step 5, Table~\ref{tab:framework_effect} compares scores on tau and $\text{tau}^*$. For high-performing models, the difference is small. However, for small models such as Qwen3.5-4B, scores drop substantially on $\text{tau}^*$, which counts empty and malformed outputs as failures. Thus, tau-bench inflates the scores of such models.

\subsection{Comparison with an MT-Centric Pipeline}
\label{sec:compare_seatau}
We compare J-tau with $\text{SEATau}_{\text{ja}}$, which machine-translates tau-bench into Japanese with the pipeline of \citet{seatau} and applies the scoring fix in Step 5, without entity adaptation, human review, or task fixes. 
In Table~\ref{tab:seatau}, an LLM judge attributes each failed trial to the agent, user simulator, or evaluation framework.
$\text{SEATau}_{\text{ja}}$ has many framework-caused failures and fewer successes, especially in retail.\footnote{This is mainly because tool descriptions are in Japanese, while tool implementations still expect English arguments.}
Without entity adaptation, user names remain in English in the database but are written in katakana in the Japanese conversation, which causes database lookup failures (see the simulation in Table~\ref{tab:seatau-kata-t47}).
Such inconsistencies induce unnatural behavior even in successful trials, making the environment unsuitable as a benchmark. 
In contrast, J-tau has fewer such failures thanks to its localization and task verification.
Appendix~\ref{app:seatau_details} provides experimental details and example simulations.

\section{Conclusion}
We presented J-tau, the first publicly available Japanese benchmark for interactive tool-using LLM agents.
Beyond machine translation, we localized tau-bench through entity adaptation and human review, and fixed its invalid tasks and a scoring flaw, enabling more accurate evaluation.
Our analysis showed that machine translation alone increases framework-caused failures.
We hope J-tau advances Japanese LLM agents and our insights benefit benchmark localization for other languages.

\section*{Limitations}
\begin{itemize}[leftmargin=*, itemsep=2pt, parsep=0pt, topsep=0pt]
    \item J-tau covers three domains (airline, retail, and telecom) and does not include the banking\_knowledge domain introduced in the latest version of tau-bench~\cite{tauknowledge}.
    \item J-tau covers only Japanese and cannot be used to evaluate agents in other languages.
    Concurrent work released within one month before our submission extends tau-bench to more languages~\citep{babelarena,tau-multilingual}, but these benchmarks may still contain the invalid tasks and the aggregation flaw we identify (Appendix~\ref{app:related_work}).
    Since our task and scoring fixes are language-independent, extending our pipeline to other languages is left for future work.
    \item Addresses and other user information in the airline and retail domains are generated with \texttt{Faker} and thus include nonexistent place names and addresses, limiting their fidelity to real-world data.
    \item Because the solvability verification in Step 4 (Section~\ref{sec:construction-pipeline}) relies only on GLM-5.2-FP8, it may reflect biases specific to this model, and a small number of invalid tasks may remain undetected.
    \item As shown in Section~\ref{sec:compare_seatau}, J-tau has more failures attributed to the user simulator than $\text{SEATau}_{\text{ja}}$, especially in the retail domain. This is likely because the user simulator follows the user simulation guideline less faithfully in Japanese than in English (see Appendix~\ref{app:seatau_user_error} for an ablation study). Since this may change as models improve, revisiting the choice of user simulator is left for future work.
    \item The original tau-bench prompt instructs models to produce outputs in JSON format. During evaluation after constructing J-tau, we found that this instruction affects some models, such as Kimi-K3~\cite{kimik3}, more strongly in Japanese, causing them to output JSON strings instead of natural-language responses to the user. Since these outputs are counted as failures, the scores of such models may be underestimated (see Appendix~\ref{app:kimi_analysis} for an ablation study). 
    We plan to remove this instruction in the revised version.
    
\end{itemize}

\section*{Ethical Considerations}
J-tau contains no personally identifiable information, as all user information, such as names, addresses, and phone numbers, is fictitious.
The human review was conducted by in-house annotators as part of their regular duties, for which they received their standard salaries.

\bibliography{main}

@inproceedings{maps,
    title = "{MAPS}: A Multilingual Benchmark for Agent Performance and Security",
    author = "Hofman, Omer  and
      Brokman, Jonathan  and
      Rachmil, Oren  and
      Bose, Shamik  and
      Pahuja, Vikas  and
      Shimizu, Toshiya  and
      Starostina, Trisha  and
      Marchisio, Kelly  and
      Goldfarb-Tarrant, Seraphina  and
      Vainshtein, Roman",
    booktitle = "Findings of the {A}ssociation for {C}omputational {L}inguistics: {EACL} 2026",
    year = "2026",
    url = "https://aclanthology.org/2026.findings-eacl.42/",
    pages = "821--845",
}

@inproceedings{
taubench,
title={{$\tau$-bench}: A Benchmark for Tool-Agent-User Interaction in Real-World Domains},
author={Shunyu Yao and Noah Shinn and Pedram Razavi and Karthik R Narasimhan},
booktitle={The Thirteenth International Conference on Learning Representations: ICLR 2025},
year={2025},
url={https://openreview.net/forum?id=roNSXZpUDN}
}

@inproceedings{
tau2bench,
title={{$\tau^2$-Bench}: Evaluating Conversational Agents in a Dual-Control Environment},
author={Victor Barres and Honghua Dong and Soham Ray and Xujie Si and Karthik R Narasimhan},
booktitle={Forty-third International Conference on Machine Learning: ICML 2026},
year={2026},
url={https://openreview.net/forum?id=OC2z7iSQKa}
}

@inproceedings{
tauknowledge,
title={{$\tau$-Knowledge}: Evaluating Conversational Agents over Unstructured Knowledge},
author={Quan Shi and Alexandra Zytek and Pedram Razavi and Karthik R Narasimhan and Victor Barres},
booktitle={Forty-third International Conference on Machine Learning: ICML 2026},
year={2026},
url={https://openreview.net/forum?id=XHZK5abtw2}
}

@inproceedings{lost_in_sim,
    title = "{{Lost in Simulation}}: {LLM}-Simulated Users are Unreliable Proxies for Human Users in Agentic Evaluations",
    author = "Seshadri, Preethi  and
      Cahyawijaya, Samuel  and
      Odumakinde, Ayomide  and
      Singh, Sameer  and
      Goldfarb-Tarrant, Seraphina",
    booktitle = "Proceedings of the 64th Annual Meeting of the {A}ssociation for {C}omputational {L}inguistics: ACL 2026",
    year = "2026",
    url = "https://aclanthology.org/2026.acl-long.2192/",
    pages = "47423--47439",
}

@article{seatau,
       title={{SEATauBench}: Progressively Adapting Tool-Agent-User Evaluation into Low-Resource {Southeast Asian} Languages},
      author={My Chiffon Nguyen and Aulia Adila and Saksorn Ruangtanusak and Kittiphat Leesombatwathana and Vissuta Gunawan Lim and Patomporn Payoungkhamdee and Samuel Cahyawijaya},
    year={2026},
      url={https://arxiv.org/abs/2606.28715}, 
      eprint={2606.28715},
    journal = {ar{X}iv preprint ar{X}iv:2606.28715},
}

@inproceedings{massive_agents,
    title = "{{MASSIVE-Agents}}: A Benchmark for Multilingual Function-Calling in 52 Languages",
    author = "Kulkarni, Mayank  and
      Mazzia, Vittorio  and
      Gaspers, Judith  and
      Hench, Chris  and
      FitzGerald, Jack",
    booktitle = "Findings of the Association for Computational Linguistics: EMNLP 2025",
    year = "2025",
    url = "https://aclanthology.org/2025.findings-emnlp.1099/",
    pages = "20193--20215",
}

@misc{gpt56sol,
      title={{GPT-5.6}: Frontier intelligence that scales with your ambition}, 
      author={OpenAI},
      year={2026},
      url={https://openai.com/index/gpt-5-6/}, 
}

@misc{claudefable,
      title={{Claude Fable 5 and Claude Mythos 5}}, 
      author={Anthropic},
      year={2026},
      url={https://www.anthropic.com/news/claude-fable-5-mythos-5}, 
}

@misc{claudeopus,
      title={{Introducing Claude Opus 4.8}}, 
      author={Anthropic},
      year={2026},
      url={https://www.anthropic.com/news/claude-opus-4-8}, 
}

@misc{gemini,
      title={{Gemini 3.1 Pro}}, 
      author={{Google DeepMind}},
      year={2026},
      url={https://deepmind.google/models/model-cards/gemini-3-1-pro/}, 
}

@article{kimik3,
      title={{Kimi K3}: Open Frontier Intelligence}, 
      author={{Kimi Team} and Tongtong Bai and Yifan Bai and Yiping Bao and M. C. and Jianfeng Cai and Xinyuan Cai and Peizhou Cao and Yuxuan Cao and Ziwei Chai and Y. Charles and H. S. Che and Guanduo Chen and Guangyu Chen and Guanzheng Chen and Huarong Chen and Jia Chen and Jianlong Chen and Jun Chen and Kexin Chen and Peng Chen and Ruijue Chen and Wentao Chen and Xin Chen and Yang Chen and Yanru Chen and Yifei Chen and Yingjiang Chen and Yuankun Chen and Yujie Chen and Yutian Chen and Zhirong Chen and Dazhi Cheng and Yean Cheng and Jialei Cui and Jingbing Cui and Anqi Dai and Jiaqi Deng and Hao Ding and Rui Ding and Shaofeng Ding and Mengfan Dong and Mengnan Dong and Yuhao Dong and Yuxin Dong and Angang Du and Chenzhuang Du and Dikang Du and Jusen Du and Yulun Du and Yu Fan and Jing Feng and Qiulin Feng and Yichen Feng and Kelin Fu and Qiang Fu and Fuxuan Gao and Hongcheng Gao and Jingyue Gao and Tong Gao and Weijia Gao and Shangyi Geng and Jie Gong and Linhu Gong and Shengao Gong and Xiaochen Gong and Qizheng Gu and Yicheng Gu and Shuhao Guan and Haiqing Guo and Shiqi Guo and Xiang Guo and Zhengyan Guo and Beixi Hao and Wenxin Hao and Xiaoru Hao and Dailan He and Haotian He and Lehan He and Qi He and Weiran He and Xinran He and Xinyi He and Yibo He and Yunjia He and Chao Hong and Tiange Hong and Hao Hu and Jiaxi Hu and Ruikun Hu and Weiming Hu and Yangyang Hu and Zhenxing Hu and Liang Hua and Jinbin Huang and Ke Huang and Ruiyuan Huang and Siying Huang and Weixiao Huang and Yan Huang and Zhengjie Huang and Zhiqi Huang and Yulong Hui and Chaobo Jia and Yutong Jiang and Zhejun Jiang and Zuoyou Jiang and Wenyi Jin and Xinyi Jin and Yu Jing and Huanjun Kong and Guokun Lai and Aidi Li and Cheng Li and Chengyuan Li and Cong Li and Fang Li and Guanyu Li and Haoyang Li and Jia Li and Junxiong Li and Lei Li and Letian Li and Lincan Li and Weihong Li and Wentao Li and Xintong Li and Yang Li and Yishen Li and Yiwei Li and Yuxiao Li and Zhaowei Li and Zhaoxi Li and Zheming Li and Zhengxiao Li and Zhiyuan Li and Jiawei Lin and Xiaohan Lin and Yibo Lin and Zichao Lin and Ziyan Lin and Bill Liu and Boxiao Liu and Chuan Liu and Liang Liu and Shaowei Liu and Shudong Liu and Shuran Liu and Tianwei Liu and Weizhou Liu and Yangyang Liu and Yanming Liu and Yibo Liu and Yipeng Liu and Zhengying Liu and Zhiheng Liu and Enzhe Lu and Haoyu Lu and Linqiang Lu and Tingzhan Lu and Zhiyuan Lu and Aotian Luo and G. Luo and Junyu Luo and Yifan Luo and B. Lyu and Wenzhou Lyu and Shaoguang Mao and Yuan Mei and Xin Men and Minqing Ni and Yixuan Niu and Siyuan Pan and Shujun Peng and Zhangyang Qi and Ruoyu Qin and ZeChao Qin and Zeyu Qin and Haiquan Qiu and Jianxin Qiu and Jiezhong Qiu and Bowen Qu and Yuhao Qu and Zeyu Shang and Youbo Shao and Han Shen and Jincheng Shi and Juanfeng Shi and Lidong Shi and Shengyuan Shi and Wingchun Siu and Pengwei Song and Xiaoxi Song and Jianlin Su and Yunfeng Su and Zhaochen Su and Lin Sui and Jingsong Sun and Junyao Sun and Shaoning Sun and Shuzhe Sun and Tongyu Sun and Yujun Sun and Yunpeng Tai and Chuning Tang and Heyi Tang and Sirui Tang and Zecheng Tang and Chaoran Tian and Rongpeng Tian and Yu Tian and Wei Tu and Chensi Wang and Chuang Wang and Chunjie Wang and Dinglu Wang and Feng Wang and Hailong Wang and Haiming Wang and Hao Wang and Hao Wang and Huaqing Wang and Hui Wang and Jiayi Wang and Jinglong Wang and Jinhong Wang and Jiuzheng Wang and Linian Wang and Shaobo Wang and Shenzhi Wang and Shuyi Wang and Si Wang and Siyuan Wang and Tianfu Wang and Wenjue Wang and Xingran Wang and Xinmei Wang and Xinyuan Wang and Xusheng Wang and Yalin Wang and Yangkun Wang and Yao Wang and Yaoyu Wang and Yejie Wang and Yiqin Wang and Yucheng Wang and Yuzhi Wang and Zhaoji Wang and Zhaowei Wang and Zhengtao Wang and Zhenhao Wang and Zhongsheng Wang and Zifan Wang and Chu Wei and Ming Wei and Shouxin Wei and Zichen Wen and Fan Wu and Haoning Wu and Rucong Wu and Wenhao Wu and Xiaoxue Wu and Yingcong Wu and Yongqi Wu and Yuxin Wu and Zijian Wu and Xinglang Xian and Chenxuan Xiang and Yuye Xiang and Bocheng Xiao and Chenjun Xiao and Xin Xiao and Jin Xie and Xiaotong Xie and Yifeng Xie and Zhe Xie and Bowei Xing and Yiming Xiong and Baosheng Xu and Boyu Xu and Jiale Xu and Jianfan Xu and Jing Xu and Jinjing Xu and L. H. Xu and Qingtao Xu and Shuyao Xu and Suting Xu and Tiantian Xu and Tianxiang Xu and Weixin Xu and Xinran Xu and Yangchuan Xu and Ye Xu and Yueni Xu and Ziyao Xu and Haonan Xue and Junjie Yan and Yaoyao Yan and Fan Yang and Guangyao Yang and Hao Yang and Junwei Yang and Ruoyu Yang and Wenjie Yang and Xiaofei Yang and Xinyu Yang and Yi Yang and Yiling Yang and Ying Yang and Yuchen Yang and Zhen Yang and Zhilin Yang and Zian Yang and Zuhao Yang and Haotian Yao and Dan Ye and Haoran Ye and Wenjie Ye and Zhanbo Ye and Bohong Yin and Haoxiang Yin and Xietong Yin and Chengzhen Yu and Haozhen Yu and Longhui Yu and Shengnan Yu and Shuying Yu and Tianxiang Yu and Enming Yuan and Mengjie Yuan and Tongtian Yue and Wei Yue and Yang Yue and Dunyuan Zha and Haobing Zhan and B. H. Zhang and Dehao Zhang and Fei Zhang and Hao Zhang and Haoyuan Zhang and Huanyu Zhang and Jiapei Zhang and Jiaxuan Zhang and Jin Zhang and Kaiyi Zhang and Miaozhen Zhang and Puqi Zhang and Qinglei Zhang and Rong Zhang and Rui Zhang and Shaoshuai Zhang and Shiyi Zhang and Xiaobin Zhang and Xiaoyun Zhang and Y. Zhang and Yangkun Zhang and Ye Zhang and Yichi Zhang and Yikun Zhang and Yizhi Zhang and Yongting Zhang and Yu Zhang and Yutao Zhang and Yutong Zhang and Zheng Zhang and Zijing Zhang and Bin Zhao and Chenguang Zhao and Feifan Zhao and Jinglun Zhao and Jinxiang Zhao and Shuai Zhao and Wenshuo Zhao and Xiangyu Zhao and Xuanle Zhao and Yikai Zhao and Zijia Zhao and Haozhi Zheng and Huabin Zheng and Ruihan Zheng and Shaojie Zheng and Tengyang Zheng and Haofeng Zhong and Lei Zhong and Longguang Zhong and M. Zhou and Qiankang Zhou and Runjie Zhou and Ruozhang Zhou and Xinyu Zhou and Yiqiao Zhou and Zaida Zhou and Jinguo Zhu and Liya Zhu and Xinhao Zhu and Yangjunfeng Zhu and Yuxuan Zhu and Zhen Zhu and Chen Zhuang and Weiyu Zhuang and Xinxing Zu},
      year={2026},
      journal = {ar{X}iv preprint ar{X}iv:2607.24653},
      url={https://arxiv.org/abs/2607.24653}, 
      eprint={2607.24653},
}

@article{glm5,
      title={{GLM-5}: from Vibe Coding to Agentic Engineering},
      author={{GLM-5-Team} and Aohan Zeng and Xin Lv and Zhenyu Hou and Zhengxiao Du and Qinkai Zheng and Bin Chen and Da Yin and Chendi Ge and Chenghua Huang and Chengxing Xie and Chenzheng Zhu and Congfeng Yin and Cunxiang Wang and Gengzheng Pan and Hao Zeng and Haoke Zhang and Haoran Wang and Huilong Chen and Jiajie Zhang and Jian Jiao and Jiaqi Guo and Jingsen Wang and Jingzhao Du and Jinzhu Wu and Kedong Wang and Lei Li and Lin Fan and Lucen Zhong and Mingdao Liu and Mingming Zhao and Pengfan Du and Qian Dong and Rui Lu and Shuang-Li and Shulin Cao and Song Liu and Ting Jiang and Xiaodong Chen and Xiaohan Zhang and Xuancheng Huang and Xuezhen Dong and Yabo Xu and Yao Wei and Yifan An and Yilin Niu and Yitong Zhu and Yuanhao Wen and Yukuo Cen and Yushi Bai and Zhongpei Qiao and Zihan Wang and Zikang Wang and Zilin Zhu and Ziqiang Liu and Zixuan Li and Bojie Wang and Bosi Wen and Can Huang and Changpeng Cai and Chao Yu and Chen Li and Chengwei Hu and Chenhui Zhang and Dan Zhang and Daoyan Lin and Dayong Yang and Di Wang and Ding Ai and Erle Zhu and Fangzhou Yi and Feiyu Chen and Guohong Wen and Hailong Sun and Haisha Zhao and Haiyi Hu and Hanchen Zhang and Hanrui Liu and Hanyu Zhang and Hao Peng and Hao Tai and Haobo Zhang and He Liu and Hongwei Wang and Hongxi Yan and Hongyu Ge and Huan Liu and Huanpeng Chu and Jia'ni Zhao and Jiachen Wang and Jiajing Zhao and Jiamin Ren and Jiapeng Wang and Jiaxin Zhang and Jiayi Gui and Jiayue Zhao and Jijie Li and Jing An and Jing Li and Jingwei Yuan and Jinhua Du and Jinxin Liu and Junkai Zhi and Junwen Duan and Kaiyue Zhou and Kangjian Wei and Ke Wang and Keyun Luo and Laiqiang Zhang and Leigang Sha and Liang Xu and Lindong Wu and Lintao Ding and Lu Chen and Minghao Li and Nianyi Lin and Pan Ta and Qiang Zou and Rongjun Song and Ruiqi Yang and Shangqing Tu and Shangtong Yang and Shaoxiang Wu and Shengyan Zhang and Shijie Li and Shuang Li and Shuyi Fan and Wei Qin and Wei Tian and Weining Zhang and Wenbo Yu and Wenjie Liang and Xiang Kuang and Xiangmeng Cheng and Xiangyang Li and Xiaoquan Yan and Xiaowei Hu and Xiaoying Ling and Xing Fan and Xingye Xia and Xinyuan Zhang and Xinze Zhang and Xirui Pan and Xu Zou and Xunkai Zhang and Yadi Liu and Yandong Wu and Yanfu Li and Yidong Wang and Yifan Zhu and Yijun Tan and Yilin Zhou and Yiming Pan and Ying Zhang and Yinpei Su and Yipeng Geng and Yong Yan and Yonglin Tan and Yuean Bi and Yuhan Shen and Yuhao Yang and Yujiang Li and Yunan Liu and Yunqing Wang and Yuntao Li and Yurong Wu and Yutao Zhang and Yuxi Duan and Yuxuan Zhang and Zezhen Liu and Zhengtao Jiang and Zhenhe Yan and Zheyu Zhang and Zhixiang Wei and Zhuo Chen and Zhuoer Feng and Zijun Yao and Ziwei Chai and Ziyuan Wang and Zuzhou Zhang and Bin Xu and Minlie Huang and Hongning Wang and Juanzi Li and Yuxiao Dong and Jie Tang},
      journal = {ar{X}iv preprint ar{X}iv:2602.15763},
      url={https://arxiv.org/abs/2602.15763}, 
      eprint={2602.15763},
      year={2026},
}

@article{gpt-oss,
      title={{gpt-oss-120b \& gpt-oss-20b Model Card}}, 
      author={OpenAI},
      year={2025},
      url={https://arxiv.org/abs/2508.10925}, 
    journal={{arXiv preprint arXiv:2508.10925}},
}

@misc{qwen3.5,
    title  = {{Qwen3.5}: Towards Native Multimodal Agents},
    author = {{Qwen Team}},
    month  = {February},
    year   = {2026},
    url    = {https://qwen.ai/blog?id=qwen3.5},
}

@misc{gemma4,
  title = {{Gemma 4}: Byte for byte, the most capable open models},
  author = {Clement Farabet and Olivier Lacombe},
  year = {2026},
  url = {https://blog.google/innovation-and-ai/technology/developers-tools/gemma-4},
}

@article{llm-jp,
    title={{LLM-jp}: A Cross-organizational Project for the Research and Development of Fully Open {Japanese LLMs}}, 
    year={2024},
    author={{LLM-jp}},
  journal={{arXiv preprint arXiv:2407.03963}},
  url={https://arxiv.org/abs/2407.03963}, 
}

@inproceedings{jamcqa,
    title = {{JamC-QA}: A Multiple-Choice Question Answering Benchmark for {Japan}-Specific Knowledge},
    author = "Oka, Teruaki  and
      Shibata, Tomohide  and
      Yoshida, Nao",
    booktitle = "Proceedings of the Fifteenth Language Resources and Evaluation Conference: LREC 2026",
    year = "2026",
    url = "https://aclanthology.org/2026.lrec-1.356/",
    pages = "4536--4546",
}

@inproceedings{jhumaneval,
    author    = {Mii Sato and Shiho Takano and Teruno Kajiura and Kimio Kuramitsu},
    title     = {Does {LLM} Cross-Lingual Knowledge Transfer Occur Through {Japanese} Continual Pre-training? (in {Japanese})},
    booktitle = {Proceedings of the 30th Annual Meeting of the Association for Natural Language Processing},
    year      = {2024},
    url       = {https://www.anlp.jp/proceedings/annual_meeting/2024/pdf_dir/P10-9.pdf},
}

@inproceedings{mifeval,
    title = {{M-IFEval}: Multilingual Instruction-Following Evaluation},
    author = "Dussolle, Antoine  and
      Carde{\~n}a, A.  and
      Sato, Shota  and
      Devine, Peter",
    booktitle = "Findings of the Association for Computational Linguistics: NAACL 2025",
    year = "2025",
    url = "https://aclanthology.org/2025.findings-naacl.344/",
    pages = "6161--6176",
}

@inproceedings{mmluprox,
    title = {{MMLU-ProX}: A Multilingual Benchmark for Advanced Large Language Model Evaluation},
    author = "Xuan, Weihao  and
      Yang, Rui  and
      Qi, Heli  and
      Zeng, Qingcheng  and
      Xiao, Yunze  and
      Feng, Aosong  and
      Liu, Dairui  and
      Xing, Yun  and
      Wang, Junjue  and
      Gao, Fan  and
      Lu, Jinghui  and
      Jiang, Yuang  and
      Li, Huitao  and
      Li, Xin  and
      Yu, Kunyu  and
      Dong, Ruihai  and
      Gu, Shangding  and
      Li, Yuekang  and
      Xie, Xiaofei  and
      Juefei-Xu, Felix  and
      Khomh, Foutse  and
      Yoshie, Osamu  and
      Chen, Qingyu  and
      Teodoro, Douglas  and
      Liu, Nan  and
      Goebel, Randy  and
      Ma, Lei  and
      Marrese-Taylor, Edison  and
      Lu, Shijian  and
      Iwasawa, Yusuke  and
      Matsuo, Yutaka  and
      Li, Irene",
    booktitle = "Proceedings of the 2025 Conference on Empirical Methods in Natural Language Processing: EMNLP 2025",
    year = "2025",
    url = "https://aclanthology.org/2025.emnlp-main.79/",
    pages = "1513--1532",
}

@inproceedings{gpqadiamond,
    title = {{BenchMAX}: A Comprehensive Multilingual Evaluation Suite for Large Language Models},
    author = "Huang, Xu  and
      Zhu, Wenhao  and
      Hu, Hanxu  and
      He, Conghui  and
      Li, Lei  and
      Huang, Shujian  and
      Yuan, Fei",
    booktitle = "Findings of the Association for Computational Linguistics: EMNLP 2025",
    year = "2025",
    url = "https://aclanthology.org/2025.findings-emnlp.909/",
    pages = "16751--16774",
}

@inproceedings{
    polymath,
    title={{PolyMath}: Evaluating Mathematical Reasoning in Multilingual Contexts},
    author={Yiming Wang and Pei Zhang and Jialong Tang and Hao-Ran Wei and Baosong Yang and Rui Wang and Chenshu Sun and Feitong Sun and Jiran Zhang and Junxuan Wu and Qiqian Cang and Yichang Zhang and Fei Huang and Junyang Lin and Fei Huang and Jingren Zhou},
    booktitle={The Thirty-ninth Annual Conference on Neural Information Processing Systems Datasets and Benchmarks Track},
    year={2026},
    url={https://openreview.net/forum?id=B1vCImy6yI}
}

@inproceedings{
bfcl,
title={The {Berkeley} Function Calling Leaderboard ({BFCL}): From Tool Use to Agentic Evaluation of Large Language Models},
author={Shishir G Patil and Huanzhi Mao and Fanjia Yan and Charlie Cheng-Jie Ji and Vishnu Suresh and Ion Stoica and Joseph E. Gonzalez},
booktitle={Forty-second International Conference on Machine Learning: ICML 2025},
year={2025},
url={https://openreview.net/forum?id=2GmDdhBdDk}
}

@article{ale,
      title={{Agents' Last Exam}}, 
      author={Yiyou Sun and Xinyang Han and Weichen Zhang and Yuanbo Pang and Tianyu Wang and Yuhan Cao and Yixiao Huang and Chris Duroiu and Haoyun Zhang and Jeffrey Lin and Weishu Zhang and Tyler Zeng and Ying Yan and Bo Liu and Hanson Wen and Mingyang Xu and Xiaoyuan Liu and Zimeng Chen and Weiyan Shi and Amanda Dsouza and Vincent Sunn Chen and Patrick Bryant and Carl Boettiger and Yamini Rangan and Bradley Rothenberg and Kyle Steinfeld and Arvind Rao and Tapio Schneider and Georgios Yannakakis and Laure Zanna and Kaan Ozbay and Ida Sim and Tarek Zohdi and George Em Karniadakis and Jack Gallant and Teresa Head-Gordon and Yushan Li and Wenxi Deng and Tao Sun and Huiqi Wang and Zhun Wang and Justin Xu and Chris Yuhao Liu and Yafei Cheng and Rongwang Hu and Aras Bacho and Shengcao Cao and Zengyi Qin and Yixiong Chen and Hengduan Fan and Hao Liu and Lin Zeng and Shashank Muralidhar Bharadwaj and Litian Gong and Yingxuan Yang and Maojia Song and Ruheng Wang and Zongzheng Zhang and Honglin Bao and Shuo Lu and Jianhong Tu and Zhonghua Wang and Zheng Zhang and Zijiao Chen and Yanqiong Jiang and Zhendong Li and Bohan Lyu and Chang Ma and Peiran Xu and Benran Zhang and Shangding Gu and Haoyue Hua and Haoyang Li and Wanzhe Liao and Chengzhi Liu and Junbo Peng and Haoran Sun and Zechen Xu and Bo Chen and Jiayi Cheng and Yi Jiang and Keying Kuang and Yuan Li and Youbang Pan and Ziyan Rao and Alexander Schubert and Yifan Shen and Vincent Siu and Xiatao Sun and Kangqi Zhang and Xiaopan Zhang and Yuchen Zhu and Ishaan Singh Chandok and Lei Ding and Jingxuan Fan and Andrew Glover and Jiaming Hu and Yiran Hu and Wenbo Huang and Zixin Jiang and Haoran Jin and Lukas Kim and Ming Liu and Yang Liu and Alireza Rafiei and Xuhuan Shen and Kunyang Sun and Sophia Sun and Ting Sun and Eric Wang and Yixin Wang and Hanwen Xing and Sihan Xu and Yuzheng Xu and Zhongxing Xu and Zhiling Yan and Boqin Yuan and Ruiqi Zhang and Yifan Zhang and Zibo Zhao and Liana and Santanu Bosu Antu and Haoyue Bai and Carlo Bosio and Joseph Cavanagh and Patricia Cavazos-Rehg and Tianxing Chen and Xuewen Chen and Yipu Chen and Chenyu Zhu and Chen Dai and Stefano De Castro and Yunfu Deng and Kaustubh Dhole and Jiayuan Ding and Chenchen Du and Zhehang Du and Hao Fan and Run-Ze Fan and Hengyu Fu and Shi Gu and Yifan Gu and Charlie Guo and Baihe Huang and Baixiang Huang and Rimika Jaiswal and Zhihan Jiang and Ran Jin and Erin Kasson and Xin Lan and Joseph Lee and Deren Lei and Chenyu Li and Daofeng Li and Haitao Li and Hongwei Li and Jingyan Li and Xiao Li and Yi Li and Yinsheng Li and Yuangang Li and Zhixu Li and Wenyu Liang and Longtai Liao and Kevin Qinghong Lin and Andy Zeyi Liu and Che Liu and Jiaming Liu and Kaiyuan Liu and Xuan Liu and Pan Lu and Wenbo Lv and Yicheng Lyu and Qiuyang Mang and Kyle Montgomery and Yuzhou Nie and Ruoxi Ning and Jorin Overwiening and Xu Pan and Layna Paraboschi and Core Francisco Park and Justin Purnomo and Swati Rajwal and Scott Rankin and Bixuan Ren and Yiren Rong and HaoYang Shang and Ventus Shaw and Fiona Shen and Jiawei Shen and Minqi Shi and Shi Qiu and Huaxiu Yao and Tianneng Shi and Jonah So and Vladislav Susoy and Hannah Szlyk and Haocheng Wang and Jialu Wang and Wei Wang and Xinyu Wang and Zehao Wang and Dowling Wong and Angela Wu and Dehao Wu and Fangyu Wu and Mengyuan "Millie" Wu and Yu Wu and Yuchen Wu and Yuhao Wu and Qingpo Wuwu and Weihang Xiao and Yongyi Xiong and Fan Xu and Ruiling Xu and Mingxuan Yan and Benjamin Yang and Jirong Yang and Sen Yang and Xiaoli Yang and Yushi Yang and Haoran Ye and Xiaohu Yu and Zhengming Yu and Chenlong Zhang and Chi Zhang and Hanning Zhang and Hanwen Zhang and Junge Zhang and Kunpeng Zhang and Song Zhang and Wenjin Zhang and Wenshuo Zhang and Ying Zhang and Yizhi Zhang and Brian Zhao and Qijian Zhao and Yimin Zhao and Yuhaohua Zheng and Liwei Zhou and Tianyue Zhou and Sichen Zhu and Siqi Zhu and Yan Zhu and Yishu Zhu and Jierui Zuo and Chonghao Cai and Helena Casademunt and Wenjia Chen and Cheng Cheng and Nawen Deng and Rao Fu and Tianfu Fu and Yifan Han and He Ren and Zhenyu He and Qiao Jin and Langlang Li and Yuetai Li and Sylvia Liu and Lu Lu and Luqing Zhou and Subhabrata Mukherjee and Yunqi Ouyang and Yin Ren and Dawei Shi and Haoran Wu and Zhiyue Wu and Hannah Yao and Zhuoran Yi and Jenny Yu and Rhea Zhan and Hang Zhou and Blake Zhu and Junfan Zhu and Alan Yuille and Yang Liu and Russell Alan Poldrack and Jiachen Li and Zhenglu Li and Molei Tao and Jing Huang and Wenqi Shi and Costas Spanos and Lichao Sun and Chenguang Wang and Orson Xu and Zhen Dong and Hector Gomez and Aylin Caliskan and Ali Emami and Haimin Hu and Zhi Li and Lihui Liu and Murphy Niu and Yi Shao and Jianxin Sun and Mikko Tolonen and Ting Wang and Sanjiv Das and Yanjun Gao and Wenbo Guo and Erika J Schneider and Zhiyong Lu and Yian Ma and Mark Mueller and Radha Poovendran and Somayeh Sojoudi and Yinglun Zhu and Dawn Song},
      year={2026},
  journal={{arXiv preprint arXiv:2606.05405}},
  url={https://arxiv.org/abs/2606.05405}, 
}

@inproceedings{gaia_lilt,
    title = "{GAIA}-v2-{LILT}: Multilingual Adaptation of Agent Benchmark beyond Translation",
    author = "Kim, Yunsu  and
      Uhlig, Kaden  and
      Wuebker, Joern",
    booktitle = "Proceedings of the 1st Workshop on Multilinguality in the Era of Large Language Models ({M}e{LLM} 2026)",
    year = "2026",
    url = "https://aclanthology.org/2026.mellm-1.13/",
    pages = "140--148",
}

@inproceedings{
    swebench,
    title={{SWE}-bench: Can Language Models Resolve Real-world {GitHub} Issues?},
    author={Carlos E Jimenez and John Yang and Alexander Wettig and Shunyu Yao and Kexin Pei and Ofir Press and Karthik R Narasimhan},
    booktitle={The Twelfth International Conference on Learning Representations: ICLR 2024},
    year={2024},
    url={https://openreview.net/forum?id=VTF8yNQM66}
}

@article{DBLP:journals/fcsc/WangMFZYZCTCLZWW24,
  author       = {Lei Wang and
                  Chen Ma and
                  Xueyang Feng and
                  Zeyu Zhang and
                  Hao Yang and
                  Jingsen Zhang and
                  Zhiyuan Chen and
                  Jiakai Tang and
                  Xu Chen and
                  Yankai Lin and
                  Wayne Xin Zhao and
                  Zhewei Wei and
                  Jirong Wen},
  title        = {A survey on large language model based autonomous agents},
  journal      = {Frontiers Comput. Sci.},
  volume       = {18},
  number       = {6},
  pages        = {186345},
  year         = {2024},
  url          = {https://doi.org/10.1007/s11704-024-40231-1},
  doi          = {10.1007/S11704-024-40231-1},
  bibsource    = {dblp computer science bibliography, https://dblp.org}
}

@article{tau-fix,
title={{SABER}: Small Actions, Big Errors -- Safeguarding Mutating Steps in {LLM} Agents},
      author={Alejandro Cuadron and Pengfei Yu and Yang Liu and Arpit Gupta},
      year={2025},
      url={https://arxiv.org/abs/2512.07850}, 
        journal={{arXiv preprint arXiv:2512.07850}},
}

@article{kbrowsecomp,
      title={{K-BrowseComp}: A Web Browsing Agent Benchmark Grounded in {Korean} Contexts}, 
      author={Nahyun Lee and Dongkeun Yoon and Guijin Son and Geewook Kim and Dayoon Ko and Jeonghun Park and Haneul Yoo and Jaewon Cho and Junghun Park and Changyoon Lee and Kyochul Jang and Jaeyeon Kim and Eunsu Kim and Woojin Cho and Seungone Kim},
      year={2026},
      url={https://arxiv.org/abs/2606.02404}, 
      journal={{arXiv preprint arXiv:2606.02404}},
}

@article{browsecompzh,
      title={{BrowseComp-ZH}: Benchmarking Web Browsing Ability of Large Language Models in {Chinese}}, 
      author={Peilin Zhou and Bruce Leon and Xiang Ying and Can Zhang and Yifan Shao and Qichen Ye and Dading Chong and Zhiling Jin and Chenxuan Xie and Meng Cao and Yuxin Gu and Sixin Hong and Jing Ren and Jian Chen and Chao Liu and Yining Hua},
      year={2025},
      url={https://arxiv.org/abs/2504.19314}, 
      journal={{arXiv preprint arXiv:2504.19314}},
}

@article{polyworkbench,
      title={{PolyWorkBench}: Benchmarking {LLM} Agents for Cross-Lingual Long-Horizon Workflows}, 
      author={Hongliang Li and Yijin Liu and Zhiwei Zhang and Zihe Liu and Xinyue Lou and Jinan Xu and Fandong Meng and Kaiyu Huang},
      year={2026},
      url={https://arxiv.org/abs/2607.06008}, 
      journal={{arXiv preprint arXiv:2607.06008}},
}

@article{ticketbench,
      title={{Ticket-Bench}: A Kickoff for Multilingual and Regionalized Agent Evaluation}, 
      author={Thales Sales Almeida and João Guilherme Alves Santos and Thiago Laitz and Giovana Kerche Bonás},
      year={2025},
      url={https://arxiv.org/abs/2509.14477}, 
      journal={{arXiv preprint arXiv:2509.14477}},
}

@article{omnilingualgaia2,
      title={{OmnilingualGAIA2}: Evaluating the Multilingual Gap in Frontier {AI} Agents}, 
      author={Andrea Caciolai and Pere-Lluís Huguet Cabot and Chierh Cheng and Albert Ventayol-Boada and Gabriel Mejia Gonzalez and Christophe Ropers and Lucas Bandarkar and Sebastian Ruder and Darlene Sakakihara and Elliot Yun and Pierre Andrews and Grégoire Mialon and Romain Froger and Marta R. Costa-jussà},
      year={2026},
      url={https://arxiv.org/abs/2608.08775}, 
      journal={{arXiv preprint arXiv:2608.08775}},
}

@inproceedings{mlcl,
    title = "{Lost in Execution}: On the Multilingual Robustness of Tool Calling in Large Language Models",
    author = "Luo, Zheng  and
      Kutralingam, T Pranav  and
      Okoani, Ogochukwu N.  and
      Xu, Wanpeng  and
      Wei, Hua  and
      Hu, Xiyang",
    booktitle = "Proceedings of the 64th Annual Meeting of the {A}ssociation for {C}omputational {L}inguistics: ACL 2026",
    year = "2026",
    url = "https://aclanthology.org/2026.acl-long.2039/",
    pages = "44059--44077",
}

@article{babelarena,
      title={{BabelArena}: A Large-Scale Multilingual Benchmark for {LLM} Agents}, 
      author={Peng Kuang and Yuchun Fan and Jiangnan Li and Minghao Wu and Jialong Tang and Hao-Ran Wei and Weixuan Wang and Jianhong Tu and Baosong Yang and Tong Xiao},
      year={2026},
      url={https://arxiv.org/abs/2609.23490}, 
      journal={{arXiv preprint arXiv:2609.23490}},
}

@article{qwen3,
      title={{Qwen3 Technical Report}}, 
      author={{Qwen Team}},
      year={2025},
      url={https://arxiv.org/abs/2505.09388}, 
      journal={{arXiv preprint arXiv:2505.09388}},
}

@article{tau-multilingual,
      title={{$\tau$-Multilingual}: Benchmarking Voice Agents Across Languages}, 
      author={Soham Ray and Edgard dos Santos Paiva and Ruben Valenzuela and Karthik Narasimhan and Keshav Dhandhania and Victor Barres},
      year={2026},
      url={https://arxiv.org/abs/2609.35820}, 
        journal={{arXiv preprint arXiv:2609.35820}},
}

@inproceedings{
    tauvoice,
    title={{$\tau$-Voice}: Benchmarking Full-Duplex Voice Agents on Real-World Domains},
    author={Soham Ray and Keshav Dhandhania and Victor Barres and Karthik R Narasimhan},
    booktitle={Forty-third International Conference on Machine Learning: ICML 2026},
    year={2026},
    url={https://openreview.net/forum?id=2Oj6fg0m1j}
}
\appendix

\section{Additional Related Work}
\label{app:related_work}

\paragraph{Agent benchmarks.}
LLM agent evaluation has evolved from static tasks that check single-turn function calls and their arguments~\cite{bfcl} to challenging multi-step tasks requiring advanced reasoning and environment interaction~\cite{swebench,ale}. 
Among such benchmarks, tau-bench~\cite{taubench,tau2bench,tauknowledge} evaluates agents on customer-support tasks in which they resolve the requests of an LLM-simulated user through multi-turn conversations, calling tools while complying with domain policies.

\paragraph{Multilingual agent benchmarks.}
Some studies build tasks grounded in each language and
culture~\cite{ticketbench,polyworkbench,browsecompzh,kbrowsecomp}, while others translate existing English benchmarks~\cite{maps,omnilingualgaia2,mlcl}.
The validity of the latter has been questioned: \citet{maps} report degraded agent performance in non-English languages on their machine-translated benchmarks, whereas \citet{gaia_lilt} show that much of this gap stems from translation artifacts rather than model capability.

\paragraph{Multilingual extensions of tau-bench.}
\citet{seatau} extend tau-bench to five Southeast Asian languages by progressively translating the conversations, tool definitions, and domain information, but without adapting locale-specific entities, resulting in unnatural conversations.
Concurrently, \citet{babelarena} propose a workflow that extends agent benchmarks, including tau-bench, to 23 languages including Japanese, but it does not adapt locale-specific entities, and the data have not been publicly released at the time of writing. 
\citet{tau-multilingual} extend $\tau$-Voice~\citep{tauvoice}, the voice version of tau-bench, to five languages excluding Japanese, localizing user information.
These two concurrent works verify consistency with the original tasks, but neither examines the validity of those tasks or revises the score aggregation, so the flaws described in Section~\ref{sec:taubench_limitations} may remain in their benchmarks.

\paragraph{Japanese LLM benchmarks.}
Japanese LLM benchmarks cover knowledge, reasoning, instruction following, mathematics, and coding~\cite{jamcqa,mifeval,mmluprox,gpqadiamond,polymath,jhumaneval}, but no publicly available benchmark evaluates Japanese agent capabilities in dynamic conversations with external tool calls.

\section{Details of Localization}
\subsection{Translation}
\label{app:term_dic}
Tables~\ref{app:tab:airline_dic}, \ref{app:tab:retail_dic}, and~\ref{app:tab:telecom_dic} show the domain-specific glossaries used for translation in Step~1.

\begin{table}[htbp]
    \centering
    \small
    \setlength{\tabcolsep}{1pt}       
    \setlength{\aboverulesep}{0.6pt}  
    \setlength{\belowrulesep}{0.6pt}  
    \begin{tabular}{lll}
        \toprule
        Category & en & ja \\
        \midrule
        \multirow{4}{*}{\makecell[l]{Cabin\\Class}} & cabin class & 座席クラス \\
         & business & ビジネス \\
         & economy & エコノミー \\
         & basic economy & ベーシックエコノミー \\
        \midrule
        \multirow{2}{*}{\makecell[l]{Flight\\Type}} & one way & 片道 \\
         & round trip & 往復 \\
        \midrule
        \multirow{3}{*}{\makecell[l]{Membership\\Tier}} & gold & ゴールド会員 \\
         & silver & シルバー会員 \\
         & regular & 一般会員 \\
        \midrule
        \multirow{3}{*}{\makecell[l]{Payment\\Method}} & credit\_card & クレジットカード \\
         & gift\_card & ギフトカード \\
         & certificate & 旅行クーポン \\
        \midrule
        \multirow{6}{*}{\makecell[l]{Flight\\Status}} & available & 予約可能 \\
         & delayed & 遅延 \\
         & on time & 定刻 \\
         & flying & 飛行中 \\
         & landed & 着陸済 \\
         & cancelled & キャンセル済 \\
        \bottomrule
    \end{tabular}
    \caption{Glossary for the airline domain.}
    \label{app:tab:airline_dic}
\end{table}

\begin{table}[h]
    \centering
    \small
    \setlength{\tabcolsep}{3pt}       
    \setlength{\aboverulesep}{0.6pt}  
    \setlength{\belowrulesep}{0.6pt}  
    \begin{tabular}{lll}
        \toprule
        Category & en & ja \\
        \midrule
        \multirow{7}{*}{\makecell{Order\\Status}} & pending & 保留中 \\
         & processed & 処理済み \\
         & \makecell{pending\\(item modified)}  & \makecell{保留中\\（アイテム変更済み）} \\
         & delivered & 配達済み \\
         & cancelled & キャンセル済み \\
         & exchange requested & 交換リクエスト済み \\
         & return requested & 返品リクエスト済み \\
        \midrule
        \multirow{2}{*}{\makecell{Cancellation\\Reason}} & no longer needed & 不要になった \\
         & ordered by mistake & 間違えて注文した \\
        \midrule
        \multirow{3}{*}{\makecell{Payment\\Method}} & credit\_card & クレジットカード \\
         & gift\_card & ギフトカード \\
         & paypal & PayPal \\
        \bottomrule
    \end{tabular}
    \caption{Glossary for the retail domain.}
    \label{app:tab:retail_dic}
\end{table}

\begin{table}[htbp]
    \centering
    \small
    \setlength{\tabcolsep}{3pt}       
    \setlength{\aboverulesep}{0.6pt}  
    \setlength{\belowrulesep}{0.6pt}  
    \begin{tabular}{llll}
        \toprule
        US Code & US Name & JP Code & JP Name \\
        \midrule
        JFK & New York      & HND & 羽田 (Haneda) \\
        EWR & Newark        & NRT & 成田 (Narita) \\
        LGA & LaGuardia     & IBR & 茨城 (Ibaraki) \\
        SFO & San Francisco & ITM & 伊丹 (Itami) \\
        LAX & Los Angeles   & KIX & 関西 (Kansai) \\
        DFW & Dallas        & FUK & 福岡 (Fukuoka) \\
        IAH & Houston       & KKJ & 北九州 (Kitakyushu) \\
        CLT & Charlotte     & HSG & 佐賀 (Saga) \\
        MIA & Miami         & CTS & 新千歳 (New Chitose) \\
        MCO & Orlando       & OKD & 丘珠 (Okadama) \\
        ATL & Atlanta       & OKA & 那覇 (Naha) \\
        ORD & Chicago       & NGO & 中部 (Chubu) \\
        SEA & Seattle       & SDJ & 仙台 (Sendai) \\
        LAS & Las Vegas     & HIJ & 広島 (Hiroshima) \\
        PHX & Phoenix       & KMJ & 熊本 (Kumamoto) \\
        DEN & Denver        & MYJ & 松山 (Matsuyama) \\
        MSP & Minneapolis   & KOJ & 鹿児島 (Kagoshima) \\
        DTW & Detroit       & FSZ & 静岡 (Shizuoka) \\
        BOS & Boston        & UKB & 神戸 (Kobe) \\
        PHL & Philadelphia  & NGS & 長崎 (Nagasaki) \\
        \bottomrule
    \end{tabular}
    \caption{Airport mapping for the airline domain.}
    \label{app:tab:airport_dic}
\end{table}

\begin{table*}[!tp]
    \centering
    \small
    \setlength{\tabcolsep}{3pt}
    \setlength{\aboverulesep}{0.6pt}
    \setlength{\belowrulesep}{0.6pt}
    \begin{tabularx}{0.49\textwidth}[t]{@{}lLL@{}}
        \toprule
        Category & en & ja \\
        \midrule
        \multirow{4}{*}{\makecell[l]{Account\\Status}} & Active & 有効 \\
         & Suspended & 利用停止 \\
         & Pending Verification & 確認待ち \\
         & Closed & 解約済み \\
        \midrule
        \multirow{4}{*}{\makecell[l]{Line\\Status}} & Active & 有効 \\
         & Suspended & 利用停止 \\
         & Pending Activation & 有効化待ち \\
         & Closed & 解約済み \\
        \midrule
        \multirow{6}{*}{\makecell[l]{Invoice\\Status}} & Draft & 下書き \\
         & Issued & 発行済み \\
         & Awaiting Payment & 支払い待ち \\
         & Paid & 支払い済み \\
         & Overdue & 延滞 \\
         & Disputed & 係争中 \\
        \midrule
        \multirow{5}{*}{\makecell[l]{Device\\Type}} & phone & スマートフォン \\
         & router & ルーター \\
         & tablet & タブレット \\
         & watch & スマートウォッチ \\
         & other & その他 \\
        \midrule
        \multirow{10}{*}{\makecell[l]{Status Bar\\Display}} & Airplane Mode & 機内モード \\
         & No Signal & 圏外 \\
         & Poor & 弱い \\
         & Fair & やや弱い \\
         & Good & 良好 \\
         & Excellent & 非常に良好 \\
         & Data Enabled & データ有効 \\
         & Data Disabled & データ無効 \\
         & Data Saver & データセーバー \\
         & VPN Connected & VPN接続済み \\
        \midrule
        \multirow{5}{*}{\makecell[l]{Signal\\Strength}} & none & 圏外 \\
         & poor & 弱い \\
         & fair & やや弱い \\
         & good & 良好 \\
         & excellent & 非常に良好 \\
        \bottomrule
    \end{tabularx}\hfill
    \begin{tabularx}{0.49\textwidth}[t]{@{}lLL@{}}
        \toprule
        Category & en & ja \\
        \midrule
        \multirow{6}{*}{\makecell[l]{Speed Test\\Result}} & unknown & 不明 \\
         & poor & 遅い \\
         & fair & やや遅い \\
         & good & 速い \\
         & excellent & 非常に速い \\
         & no connection & 接続なし \\
        \midrule
        \multirow{4}{*}{\makecell[l]{SIM\\Status}} & active & アクティブ \\
         & missing & 未挿入 \\
         & locked with a PIN & PINロック中 \\
         & locked with a PUK & PUKロック中 \\
        \midrule
        \multirow{4}{*}{\makecell[l]{APN\\Settings}} & internet & インターネット \\
         & broken & 不正 \\
         & Incorrect & 不正 \\
         & Not Set & 未設定 \\
        \midrule
        \multirow{4}{*}{\makecell[l]{Network\\Connection\\Status}} & connected & 接続済み \\
         & searching & 検索中 \\
         & no\_service & サービスなし \\
         & emergency\_only & 緊急通話のみ \\
        \midrule
        \multirow{6}{*}{\makecell[l]{Terminology\\Consistency}} & cellular service / cellular network & モバイル通信 \\
         & cellular data & モバイルデータ \\
         & cellular connection & モバイル接続 \\
         & network technology & モバイル通信の種類 \\
         & streaming & 動画の再生 / 動画再生 \\
         & general agent policy / general policy & 一般ポリシー \\
        \midrule
        \multirow{4}{*}{\makecell[l]{Output\\Label}} & Network Mode Preference & 通信モード設定 \\
         & Failed to set network mode & 通信モードの設定に失敗しました \\
         & Network mode set to & 通信モードを設定しました \\
         & Restarting network services & 通信サービスを再起動しています \\
        \midrule
        \multirow{2}{*}{\makecell[l]{Payment\\Method}} & Credit Card & クレジットカード \\
         & Debit Card & デビットカード \\
        \midrule
        \multirow{2}{*}{\makecell[l]{App Display\\Name}} & messaging & メッセージ \\
         & browser & ブラウザ \\
        \bottomrule
    \end{tabularx}
    \caption{Glossary for the telecom domain.}
    \label{app:tab:telecom_dic}
\end{table*}

\begin{table*}[hbtp]
\centering
\small
    \setlength{\tabcolsep}{3pt}       
\begin{NiceTabular}[width=\textwidth]{X[1,l] X[7,l]}
\toprule
\textbf{Task ID} & \textbf{Modification} \\
\midrule
\multicolumn{2}{l}{\textbf{airline}} \\
\midrule
6, 7 & The user's full name in kanji cannot be inferred from a romanized user ID or email address. Added the fixed sentence ``あなたの名前は\{family name\} \{given name\}です。'' (``Your name is \{family name\} \{given name\}.'') to the user prompts of tasks that do not mention the user's name. \\
\midrule
15 & The original prompt stated, ``Since you live in Princeton, so EWR and PHL are equally convenient for you and you want to consider both.'' After the airports were localized, no suitable location existed midway between the corresponding airports. Since this does not change the task difficulty, removed the reference to the user's place of residence. \\
\midrule
24 & To match the airport localization, changed the user's preferred destination from ``the West Coast'' to ``the Tohoku region.'' \\
\midrule
39 & This task assumes a user whose utterances in the evaluation language are imperfect because it is not their first language. Changed the user's first language from French to Chinese, whose writing system is similar to that of the evaluation language. \\
\midrule
\multicolumn{2}{l}{\textbf{retail}} \\
\midrule
5--14, 35, 39, 45--48, 56--58, 67--69 & The user's full name in kanji cannot be inferred from a romanized user ID or email address. Added the fixed sentence ``あなたの名前は\{family name\} \{given name\}です。'' (``Your name is \{family name\} \{given name\}.'') to the user prompts of tasks that do not mention the user's name. \\
\bottomrule
\end{NiceTabular}
\caption{Tasks modified due to localization.}
\label{app:tab:fixed_l}
\end{table*}

\begin{table*}[hbtp]
\centering
\small
    \setlength{\tabcolsep}{3pt}      
\begin{NiceTabular}[width=\textwidth]{X[1,l] X[7,l]}
\toprule
\textbf{Task ID} & \textbf{Modification} \\
\midrule
airline \\
\midrule
7 & The total cost of upcoming flights incorrectly included already-cancelled flights. Corrected it to the correct amount of \$708 (￥106,200). \\
\midrule
29 & Specified the payment method so that the Gold Action is uniquely determined. \\
\midrule
32 & Specified the payment and refund methods so that the Gold Action is uniquely determined. \\
\midrule
33 & Specified the refund method so that the Gold Action is uniquely determined. \\
\midrule
39 & Restricted the cancellation reason to one covered by travel insurance. \\
\midrule
44 & The expected actions incorrectly included a flight taking 7.5 hours including layovers, despite the condition ``a flight of 3 hours or less including layover time.'' Corrected the actions accordingly. Also specified the payment method so that the Gold Action is uniquely determined. \\
\midrule
retail \\
\midrule
18, 91, 107 & The task required the agent to refuse an exchange for the same item, a behavior not stated in the policy. Modified the task to allow the exchange. \\
\midrule
25 / 26 & An action required to complete Task 25 was included in Task 26. Removed the action from Task 26 and moved it to Task 25. \\
\midrule
28 & Corrected the incorrect price of the garden hose. \\
\midrule
38 & Specified the cancellation reason so that the Gold Action is uniquely determined. \\
\midrule
64 & Added conditions on the item to be exchanged so that the Gold Action is uniquely determined. \\
\midrule
99 & Revised the Gold Action to match the task instructions. \\
\midrule
104 & Corrected an incorrect ID in the Gold Action. \\
\midrule
105, 110 & Corrected the Gold Action, which contradicted the DB state and required an infeasible exchange. \\
\bottomrule
\end{NiceTabular}
\caption{Tasks from the original tau-bench repaired to make them solvable.}
\label{app:tab:fixed_u}
\end{table*}

\subsection{Entity Adaptation}
\label{app:detail_entity}
In the entity adaptation of Step~2, we create a mapping from the original entities in the user information, such as personal names, addresses, phone numbers, and user IDs, to fictitious entities suited to Japanese culture.
For the retail domain, we build a dictionary of product names and product variants and replace them accordingly to ensure consistent translations.
For the airline domain, we replace the airports with domestic Japanese airports to keep the geographic setting consistent with the conversation language.
Table~\ref{app:tab:airport_dic} shows the airport mapping.
We convert monetary amounts into Japanese yen at 1 USD = 150 JPY and round them to the nearest yen, since fractional yen are not used in everyday transactions.
We also change the time zone from EST to JST.
Because the agent passes Japanese values to the tools in J-tau, we apply this localization not only to the surface text but also to DB values, enumerated values of tool arguments, and values accepted by the tool implementations, consistently converting them into Japanese according to the glossaries and mappings.

\subsection{Human Review}
\label{app:detail_annotation}
The human review checks overall translation quality and verifies that technical terms and entities are translated correctly.
\paragraph{Airline and retail.}
For the airline and retail domains, three in-house annotators specializing in Japanese--English translation review the data.
In addition to the domain policies, the annotators review task-specific text, such as the task descriptions and the user instructions given to the user simulator.
The review covers not only translation quality but also whether technical terms follow the glossaries and whether airport names and currencies are properly localized.
Because the domain policies are particularly important, all three annotators review them.

\paragraph{Telecom.}
The telecom domain involves complex technical terminology, and naive translation may produce terms that differ from those used in practice.
Therefore, three native Japanese in-house annotators with customer-support experience in the telecom industry review the Japanese translations of the domain policy, technical support manual, and workflows provided to the agent.

\subsection{Fix Tasks}
\label{app:fixed_task}
Task fixing consists of two stages: addressing issues caused by localization and repairing invalid tasks in the original tau-bench.

First, we identify tasks that become unsolvable or whose characteristics, such as difficulty, change substantially through localization.
To this end, we manually inspect the task instructions and evaluation criteria of all tasks.
Table~\ref{app:tab:fixed_l} lists the modified tasks.

Next, to find invalid tasks in the original tau-bench, we run each task three times with GLM-5.2-FP8 and examine its behavior, focusing on tasks that fail two or more times.
Table~\ref{app:tab:fixed_u} lists the repaired tasks.

\begin{table}[h]
  \centering
  \small
  \setlength{\tabcolsep}{3pt}
  \begin{tabular}{@{}lccccc@{}}
\toprule
 & tau & $\text{tau}^*$ & J-tau & $\text{J-tau}_{\text{en-tool}}$ & $\text{SEATau}_{\text{ja}}$ \\
\midrule
Conversation & EN & EN & JA & JA & JA \\
Domain policies & EN & EN & JA & JA$^\dagger$ & JA \\
\textit{Tools} & & & & & \\
\quad Names & EN & EN & EN & EN & EN \\
\quad Descriptions & EN & EN & JA & EN & JA \\
\quad Argument values & EN & EN & JA & EN & EN \\
\quad Return values & EN & EN & JA & EN$^\ddagger$ & JA \\
\quad Error messages & EN & EN & JA & EN & EN \\
Entities & US & US & JP & JP & US \\
\midrule
Fix Tasks & -- & -- & \checkmark & \checkmark & -- \\
Fix Scoring & -- & \checkmark & \checkmark & \checkmark & \checkmark \\
\bottomrule
  \end{tabular}
  \caption{Settings of the compared benchmarks. The upper part shows the language of each component, and the lower part shows whether each fix is applied. Entities refers to locale-specific entities such as personal names, addresses, currencies, and time zones. $^\dagger$Tool-related terms are kept in English. $^\ddagger$Entities in return values remain in Japanese.}
  \label{app:tab:component_lang}
  \end{table}
  \begin{table}[h]
  \setlength{\tabcolsep}{4pt}
  \small
  \centering
  \begin{tabular}{lrrrrr}
    \toprule
    Domain & Total & Unmod. & Rep.\ only & Both & Loc.\ only \\
    \midrule
    airline & 50  & 41  & 4  & 2 & 3  \\
    retail  & 114 & 80  & 12 & 0 & 22 \\
    telecom & 114 & 114 & 0  & 0 & 0  \\
    \bottomrule
  \end{tabular}
  \caption{Breakdown of the number of tasks in J-tau. Unmod.\ denotes unmodified tasks, Rep.\ only denotes tasks with only repairs of invalid tasks in the original tau-bench (Table~\ref{app:tab:fixed_u}), Loc.\ only denotes tasks with only modifications due to localization (Table~\ref{app:tab:fixed_l}), and Both denotes tasks with both.}
  \label{app:tab:task_stats}
\end{table}

\section{Settings of the Compared Benchmarks}
\label{app:bench_settings}
Table~\ref{app:tab:component_lang} summarizes the settings of the benchmarks used in this paper.
tau is the original tau-bench, and $\text{tau}^*$ is a variant of tau-bench to which only the scoring fix in Step~5 is applied (used in Section~\ref{sec:tau*}).
J-tau is our proposed benchmark, and $\text{J-tau}_{\text{en-tool}}$ is a variant of J-tau that keeps the tool-related components in English (used in Appendix~\ref{app:tool_en}).
$\text{J-tau}_{\text{en-tool}}$ reflects a setting common in real-world applications, where only the conversation is in Japanese while the tool descriptions, argument values, return values, and error messages remain in English.
$\text{SEATau}_{\text{ja}}$ is a Japanese version constructed with the SEATau pipeline (used in Section~\ref{sec:compare_seatau}).

\section{Statistics of J-tau}
\label{app:stats_jtau}
Table~\ref{app:tab:task_stats} shows the number of tasks in J-tau for each domain and modification status.
Unmod.\ denotes unmodified tasks.
Rep.\ only denotes tasks with only the repairs of invalid tasks in the original tau-bench (Table~\ref{app:tab:fixed_u}), Loc.\ only denotes tasks with only the modifications due to localization (Table~\ref{app:tab:fixed_l}), and Both denotes tasks with both.
Total denotes the total number of tasks in each domain, which equals the sum of these four categories.
Note that in Table~\ref{tab:lang_effect}, the Repaired group consists of the Rep.\ only and Both tasks, and the Unmodified group corresponds to the Unmod.\ tasks.

\section{Details of the Evaluated Models}
\label{app:models}
For inference, we use the default \texttt{generation\_config} provided in each model's official repository for open-weight models and the default API settings for proprietary models.
We set the reasoning effort to high only for the gpt-oss series and use the default for the other models.
We evaluate the following models:
\paragraph{\textbf{Proprietary Models}}
\begin{itemize}[noitemsep]
    \item \textbf{GPT-5.6 Sol}~\cite{gpt56sol}
    \item \textbf{Gemini 3.1 Pro}~\cite{gemini}
    \item \textbf{Claude Opus 4.8}~\cite{claudeopus}
    \item \textbf{Claude Fable 5}~\cite{claudefable}
\end{itemize}
\textbf{Open-Weight Models}
\begin{itemize}[noitemsep,leftmargin=*]
    \item \textbf{Kimi-K3 Series}~\cite{kimik3}
    \begin{itemize}[noitemsep]
        \item Kimi-K3 (\url{https://huggingface.co/moonshotai/Kimi-K3})
    \end{itemize}
    \item \textbf{GLM-5.2 Series}~\cite{glm5}
    \begin{itemize}[noitemsep]
        \item GLM-5.2-FP8 (\url{https://huggingface.co/zai-org/GLM-5.2-FP8})
    \end{itemize}
    \item \textbf{gpt-oss Series}~\cite{gpt-oss}
    \begin{itemize}[noitemsep]
        \item gpt-oss-120b (\url{https://huggingface.co/openai/gpt-oss-120b})
        \item gpt-oss-20b (\url{https://huggingface.co/openai/gpt-oss-20b})
    \end{itemize}
    \item \textbf{Qwen3.5 Series}~\cite{qwen3.5}
    \begin{itemize}[noitemsep]
        \item Qwen3.5-397B-A17B-FP8 (\url{https://huggingface.co/Qwen/Qwen3.5-397B-A17B-FP8})
        \item Qwen3.5-27B (\url{https://huggingface.co/Qwen/Qwen3.5-27B})
        \item Qwen3.5-9B (\url{https://huggingface.co/Qwen/Qwen3.5-9B})
        \item Qwen3.5-4B (\url{https://huggingface.co/Qwen/Qwen3.5-4B})
    \end{itemize}
    \item \textbf{gemma-4 Series}~\cite{gemma4}
    \begin{itemize}[noitemsep]
        \item gemma-4-31b-it (\url{https://huggingface.co/google/gemma-4-31b-it})
        \item gemma-4-12b-it (\url{https://huggingface.co/google/gemma-4-12b-it})
        \item gemma-4-E4B-it (\url{https://huggingface.co/google/gemma-4-E4B-it})
        \item gemma-4-E2B-it (\url{https://huggingface.co/google/gemma-4-E2B-it})
    \end{itemize}
    \item \textbf{llm-jp 4.1 Series}~\cite{llm-jp}
    \begin{itemize}[noitemsep]
    \item llm-jp-4.1-33b-thinking (\url{https://huggingface.co/llm-jp/llm-jp-4.1-33b-thinking})
    \item llm-jp-4.1-8b-thinking (\url{https://huggingface.co/llm-jp/llm-jp-4.1-8b-thinking})
    \end{itemize}
\end{itemize}

\section{Example Simulations on J-tau}
\label{app:case_study}
Tables~\ref{tab:jtau-sim15}--\ref{tab:jtau-sim36} show successful and failed conversations of GLM-5.2-FP8 in each domain, and Tables~\ref{tab:gemma2b-sim13}--\ref{tab:jtau-sim13} show failed conversations of gemma-4-E2B-it and Qwen3.5-4B in each domain.
Unnatural Japanese is shown in \textcolor{blue}{blue}, and code-switched parts are shown in \textcolor{red}{red}.

\providecommand{\agent}[1]{#1}
\providecommand{\user}[1]{#1}
\providecommand{\toolcall}[1]{#1}
\providecommand{\agenttoolcall}[1]{#1}
\providecommand{\toolresp}[1]{\textcolor{gray}{#1}}
\providecommand{\foreign}[1]{\textcolor{red}{#1}}
\providecommand{\eng}[1]{（\textit{#1}）}
\providecommand{\broken}[1]{\textcolor{blue}{#1}}
\providecommand{\codeswitch}[1]{\textcolor{red}{#1}}
\providecommand{\omitbar}{\rule[-0.3ex]{1.8em}{0.4pt}}
\providecommand{\omittext}[2]{\scriptsize\textcolor{gray}{\textit{\omitbar\hspace{0.5em}\ifnum#1=#2 row #1 omitted\else rows #1--#2 omitted\fi\hspace{0.5em}\omitbar}}}

\begin{table*}[tp]
\footnotesize
\centering
\begin{tabular}{@{}clp{0.80\textwidth}@{}}
  \toprule
  \# & Speaker & Content \\
  \midrule
    1 & \textbf{Agent} & \agent{こんにちは！本日はどのようなご用件でしょうか？\eng{Hello! How may I help you today?}} \\
    2 & \textbf{User} & \user{こんにちは。注文番号W9300146でデスクランプを注文したんですが、別のランプに変更したいんです。今購入できる一番安いデスクランプに変えたいんですが、できますか？\eng{Hello. I ordered a desk lamp under order number W9300146, but I would like to change it to a different lamp. I would like to switch to the cheapest desk lamp currently available. Is that possible?}} \\
    \multicolumn{3}{c}{\omittext{3}{8}} \\
    9 & \textbf{Agent} & \agent{ご注文の詳細を確認しました。\eng{I have checked the details of your order.}\newline \newline \textbf{現在のご注文内容：}\eng{Current order details:}\newline - 注文番号：\#W9300146\eng{Order number: \#W9300146}\newline - 商品：デスクランプ（色：ブラック、明るさ：低、電源：USB）\eng{Item: desk lamp (color: black, brightness: low, power: USB)}\newline - 価格：22,984円\eng{Price: 22,984 yen}\newline - ステータス：保留中\eng{Status: pending}\newline - お支払い：ギフトカード\eng{Payment: gift card}\newline \newline ご注文は保留中ですので、商品の変更が可能です。\eng{Your order is still pending, so the item can be changed.}\newline \newline それでは、デスクランプのバリエーション一覧を確認し、一番安いものをお探ししますね。\eng{Now I will check the list of desk lamp variants and look for the cheapest one.}} \\
    10 & \textbf{Tool} & \toolcall{[Tool call: \texttt{get\_product\_details}(product\_id="6817146515")]} \newline \toolresp{\hspace*{1.2em}\{"name": "デスクランプ", "product\_id": "6817146515", "variants": \{"9083642334": \{"item\_id": "9083642334", "options": \{"color": "ホワイト", "brightness": "高", "power source": "USB"\}, "available": true, "price": 24642.0\}, …, 
    "5320792178": \{"item\_id": "5320792178", "options": \{"color": "ブラック", "brightness": "中", "power source": "ACアダプター"\}, "available": true, "price": 20286.0\},…}\\
    11 & \textbf{Agent} & \agent{デスクランプのバリエーションを確認したところ、現在購入可能な中で一番安いのは以下の商品です：\eng{After checking the desk lamp variants, the cheapest one currently available for purchase is the following:}\newline \newline \textbf{変更先の商品：}\eng{New item:}\newline - デスクランプ（色：ブラック、明るさ：中、電源：ACアダプター）\eng{Desk lamp (color: black, brightness: medium, power: AC adapter)}\newline - 価格：20,286円\eng{Price: 20,286 yen}\newline \newline \textbf{現在の商品：}\eng{Current item:}\newline - デスクランプ（色：ブラック、明るさ：低、電源：USB）\eng{Desk lamp (color: black, brightness: low, power: USB)}\newline - 価格：22,984円\eng{Price: 22,984 yen}\newline \newline \textbf{価格差：2,698円の返金}（お支払いに使用されたギフトカードに返金されます）\eng{Price difference: a 2,698-yen refund (to the gift card used for payment)}…}
    \\
    \multicolumn{3}{c}{\omittext{12}{16}} \\
\bottomrule
\end{tabular}
  \caption{Success case of GLM-5.2-FP8 in the retail domain. The agent changes a pending order to the cheapest available desk lamp variant and refunds the 2,698-yen difference in Japanese yen.}\label{tab:jtau-sim15}
\end{table*}

\begin{table*}[tp]
\footnotesize
\centering
\begin{tabular}{@{}clp{0.80\textwidth}@{}}
  \toprule
  \# & Speaker & Content \\
  \midrule
    1 & \textbf{Agent} & \agent{こんにちは！本日はどのようなご用件でしょうか？\eng{Hello! How may I help you today?}} \\
    2 & \textbf{User} & \user{こんにちは。先日注文したワイヤレスイヤホンについて交換をお願いしたいのですが。\eng{Hello. I would like to request an exchange for the wireless earphones I ordered the other day.}\newline \newline 誤ってIPX7防水レベルのものを注文してしまいまして、自分にはその機能は不要なんです。同じ注文の中にある別のイヤホンに交換できないかと思っています。\eng{I accidentally ordered the one with IPX7 water resistance, but I do not need that feature. I am wondering whether I can exchange it for a different pair of earphones in the same order.}} \\
    \multicolumn{3}{c}{\omittext{3}{14}} \\
    15 & \textbf{Agent} & \agent{ご注文を確認いたしました。ワイヤレスイヤホンを含む注文 \#W3470184 が配達済みであることを確認しました。\eng{I have checked your order. I confirmed that order \#W3470184, which contains the wireless earphones, has been delivered.}…
    \newline \newline 
    また、交換先のイヤホンのオプションについてどのようなご希望があるかお伺いさせてください。現在の製品のバリエーションを確認いたしますので、少々お待ちください。\eng{Also, let me ask what options you would like for the replacement earphones. I will check the current product variants, so please wait a moment.}} \\
    16 & \textbf{Tool} & \toolcall{[Tool call: \texttt{get\_product\_details}(product\_id="9924732112")]} \newline \toolresp{\hspace*{1.2em}\{"name": "ワイヤレスイヤホン", "product\_id": "9924732112", "variants": \{"9580569596":…}\\
    17 & \textbf{Agent} & \agent{お客様が交換をご希望のイヤホン（IPX7防水レベル、ブルー、4時間、38,845円）を確認しました。\eng{I have identified the earphones you would like to exchange (IPX7 water resistance, blue, 4-hour battery, 38,845 yen).}\newline \newline この製品のバリエーションで現在在庫のあるものは以下の通りです：\eng{The variants of this product currently in stock are as follows:}…}\\
    \multicolumn{3}{c}{\omittext{18}{24}} \\
\bottomrule
\end{tabular}
  \caption{Failure case of GLM-5.2-FP8 in the retail domain. The user asks for a replacement from among the items in their own order, but the agent searches across all product variants instead.}\label{tab:jtau-sim32}
  \end{table*}

\begin{table*}[tp]
\footnotesize
\centering
\begin{tabular}{@{}clp{0.80\textwidth}@{}}
  \toprule
  \# & Speaker & Content \\
  \midrule
    \multicolumn{3}{c}{\omittext{1}{10}} \\
    11 & \textbf{User} & \user{…
    それから、もう一つお願いがあるんですが、現在予約に入っている受託手荷物を削除して、その分の払い戻しもお願いできないでしょうか？\eng{Also, I have one more request: could you remove the checked baggage currently included in the reservation and refund that portion as well?}\newline \newline 払い戻し先のギフトカードについては、こちらで構いません。\eng{The gift card you mentioned as the refund destination is fine with me.}} \\
    12 & \textbf{Agent} & \agent{ご希望のフライト変更について承知いたしました。ただし、受託手荷物の削除につきまして確認させてください。\eng{I understand your requested flight change. However, let me clarify the removal of the checked baggage.}\newline \newline \textbf{受託手荷物の削除について：}\eng{About removing the checked baggage:}\newline 申し訳ありませんが、ポリシー上、\textbf{予約後に受託手荷物を削除することはできない}ことになっております。受託手荷物の追加は可能ですが、削除は承ることができません。\eng{We are sorry, but our policy states that checked baggage cannot be removed after the reservation is made. Adding checked baggage is possible, but removal cannot be accepted.}\newline \newline なお、現在の予約内容を確認しますと、受託手荷物は合計1個で、お客様はシルバー会員のため、エコノミークラスの搭乗者1名につき2個まで無料です。つまり、現在の1個は無料手荷物の範囲内となっており、追加料金は発生しておりません。\eng{Your current reservation includes one checked bag in total, and as a Silver member you may check in up to two bags free of charge per economy passenger. In other words, the current one bag is within the free allowance and no extra charge applies.}…}\\
    \multicolumn{3}{c}{\omittext{13}{17}} \\
\bottomrule
\end{tabular}
  \caption{Success case of GLM-5.2-FP8 in the airline domain. The agent correctly refuses the request to remove checked baggage, in accordance with the policy.}\label{tab:jtau-sim18}
  \end{table*}

\begin{table*}[tp]
\footnotesize
\centering
\begin{tabular}{@{}clp{0.80\textwidth}@{}}
  \toprule
  \# & Speaker & Content \\
  \midrule
    1 & \textbf{Agent} & \agent{こんにちは！本日はどのようなご用件でしょうか？\eng{Hello! How may I help you today?}} \\
    2 & \textbf{User} & \user{すみません、ちょっと急ぎの相談なんですけど。\eng{Sorry, I have something rather urgent to discuss.}\newline \newline 実は金銭的な事情があって、予約しているフライトの座席クラスをビジネスからエコノミーに変更したいんです。フライトの日時や搭乗者名はそのままで、座席クラスだけ下げてもらいたいんです。\eng{Actually, due to financial reasons, I would like to change the cabin class of a reserved flight from business to economy. I want to keep the flight dates and passenger names unchanged and lower only the cabin class.}\newline \newline 今ちょっと厳しい状況でして…早く対応してもらえると助かります。\eng{I am in a tight spot right now... I would appreciate it if you could handle this quickly.}} \\
    \multicolumn{3}{c}{\omittext{3}{5}} \\
    6 & \textbf{Tool} & \toolcall{[Tool call: \texttt{get\_user\_details}(user\_id="minoru\_okada\_3817")]} \newline \toolresp{\hspace*{1.2em}\{"user\_id": "minoru\_okada\_3817", "name": \{"first\_name": "稔", "last\_name": "岡田"\}, …, 
    "reservations": ["JG7FMM", "LQ940Q", "2FBBAH", "X7BYG1", "EQ1G6C", "BOH180"]\}} \\
    7 & \textbf{Agent} & \agent{岡田様、ユーザー情報を確認いたしました。お客様のアカウントに複数の予約がございます。ビジネスクラスの予約を特定するため、順番に予約内容を確認させていただきます。少々お待ちください。\eng{Okada, I have checked your user information. There are multiple reservations on your account. To identify the business-class reservations, I will check them one by one. Please wait a moment.}\newline \newline まず1つ目の予約（JG7FMM）を確認いたします。\eng{First, I will check the first reservation (JG7FMM).}} \\
    \multicolumn{3}{c}{\omittext{8}{16}} \\
    17 & \textbf{Agent} & \agent{…
    \textbf{変更完了内容:}\eng{Completed changes:}\newline \newline - \textbf{予約ID}: JG7FMM\eng{Reservation ID: JG7FMM}\newline - \textbf{座席クラス}: ビジネス → \textbf{エコノミー} ✓\eng{Cabin class: business → \textbf{economy} ✓}…
    \newline \newline ご要望のあった節約額ですが、\textbf{989,100円}が5〜7営業日以内にクレジットカードに払い戻されます。\eng{Regarding the savings you asked about, \textbf{989,100 yen} will be refunded to the credit card within 5-7 business days.}\newline \newline お急ぎの状況とのことでしたので、少しでもお役に立てれば幸いです。他にご不明な点がございましたら、お気軽にお申し付けください。\eng{Since you mentioned you were in a hurry, I hope this helps even a little. If you have any other questions, please feel free to ask.}} \\
    18 & \textbf{User} & \user{はい、確認しました。989,100円の払い戻しなんですね。わかりました、ありがとうございます。\eng{Yes, I have checked. So the refund is 989,100 yen. I see. Thank you.}\newline \newline これでかなり助かります。ほんとにありがとうございました。他には特にありません。\eng{This helps me a lot. Thank you very much. That is all for now.}\newline \newline \#\#\#STOP\#\#\#} \\
\bottomrule
\end{tabular}
    \caption{Failure case of GLM-5.2-FP8 in the airline domain. Of the user's multiple reservations, the agent downgrades only one and leaves the others unchanged.}\label{tab:jtau-sim37}
\end{table*}

\begin{table*}[tp]
\footnotesize
\centering
\begin{tabular}{@{}clp{0.80\textwidth}@{}}
  \toprule
  \# & Speaker & Content \\
  \midrule
    \multicolumn{3}{c}{\omittext{1}{29}} \\
    30 & \textbf{User} & \user{…
    それで…再起動するように言われましたけど、再起動すると…写真とか消えたりしないでしょうか？孫の写真がたくさん入っているんです…ちょっと心配で…大丈夫ですか？\eng{And... I was told to restart the phone, but if I restart it... will my photos be deleted or anything? I have lots of photos of my grandchild on it... I am a little worried... Is it safe?}} \\
    31 & \textbf{Agent} & \agent{ご安心ください！\textbf{再起動しても写真は一切消えません。}\eng{Please rest assured! \textbf{Restarting the phone will not delete any of your photos.}}
    \newline \newline …どうぞ安心して再起動を実行してください。完了しましたら、結果をお知らせくださいませ。\eng{Please go ahead and restart with confidence. When it is done, please let me know the result.}} \\
    \multicolumn{3}{c}{\omittext{32}{34}} \\
\bottomrule
\end{tabular}
    \caption{Success case of GLM-5.2-FP8 in the telecom domain. The agent resolves a complex task while reassuring the user.}\label{tab:jtau-sim30}
\end{table*}

\begin{table*}[tp]
\footnotesize
\centering
\begin{tabular}{@{}clp{0.80\textwidth}@{}}
  \toprule
  \# & Speaker & Content \\
  \midrule
    \multicolumn{3}{c}{\omittext{1}{16}} \\
    17 & \textbf{Agent} & \agent{
    支払いリクエストを送信いたしました。請求書のステータスは「支払い待ち」に変更されました。\eng{I have sent the payment request. The bill status has been changed to ``awaiting payment.''}\newline \newline 
    …3. \textbf{支払いの実行}: お客様が支払いリクエストを承認された後、私が「make\_payment」ツールを使用して実際の支払い処理を行います。\eng{3. \textbf{Execution of the payment}: after you approve the payment request, I will carry out the actual payment using the ``make\_payment'' tool.}\newline \newline 4. \textbf{回線の再開}: 支払いが完了し、請求書のステータスが「支払い済み」に更新されたことを確認した後、回線の利用停止を解除いたします。\eng{4. \textbf{Reactivation of the line}: after the payment is completed and the bill status is updated to ``paid,'' I will lift the suspension of the line.}\newline \newline お客様側で支払いリクエストの確認と承認をお願いできますでしょうか？\eng{Could you please check and approve the payment request on your side?}} \\
    18 & \textbf{User} & \user{はい、承知しました。支払いリクエストを確認します。\eng{All right, understood. I will check the payment request.}} \\
    19 & \textbf{User Tool} & \toolcall{[Tool call: \texttt{check\_payment\_request}()]} \newline \toolresp{\hspace*{1.2em}請求書 B1234321 に対する 9,750円 の支払いリクエストがあります。}\eng{(There is a payment request of 9,750 yen for invoice B1234321.)} \\
    20 & \textbf{User} & \user{支払いリクエストを確認しました。請求書B1234321に対して9,750円の請求が来ていますね。内容に間違いありませんので、支払いを進めてください。\eng{I have checked the payment request. There is a charge of 9,750 yen for bill B1234321. The details are correct, so please proceed with the payment.}} \\
    21 & \textbf{Agent} & \agent{恐れ入りますが、現在私が使用できるツールでは支払い処理を実行できない状況です。お客様の問題を確実に解決するため、有人オペレーターにおつなぎいたします。\eng{I am afraid that, with the tools currently available to me, I cannot execute the payment process. To make sure your problem is resolved, I will connect you to a human operator.}} \\
    \multicolumn{3}{c}{\omittext{22}{24}} \\
    \bottomrule
    \end{tabular}
    \caption{Failure case of GLM-5.2-FP8 in the telecom domain. The agent offers to make a payment that only the user can execute, and fails.}\label{tab:jtau-sim36}

\end{table*}

\begin{table*}[tp]
\footnotesize
\centering
\begin{tabular}{@{}clp{0.80\textwidth}@{}}
  \toprule
  \# & Speaker & Content \\
  \midrule
    \multicolumn{3}{c}{\omittext{1}{10}} \\
    11 & \textbf{Agent} & \agent{
    … \textbf{次のステップとして、フライトが「キャンセルされた」のか、「遅延した」のか、それとも「航空会社による欠航」なのか、具体的なキャンセル理由を教えていただけますでしょうか？}\eng{As the next step, could you tell me the specific cancellation reason — whether the flight was `cancelled', `delayed', or `cancelled by the airline'?}\newline \newline 理由が判明しましたら、その情報に基づき、補償の可能性について再度確認し、対応を進めます。\eng{Once the reason is clear, I will re-check the possibility of compensation based on it and proceed.}} \\
    12 & \textbf{User} & \user{フライトはキャンセルされたんです！航空会社側の都合でキャンセルされたんですから、当然補償の対象になりますよね？…\eng{The flight WAS cancelled! It was cancelled due to the airline's circumstances, so naturally I'm entitled to compensation, right?}
    } \\
    13 & \textbf{Agent} & \agent{
    … ご要望に基づき、予約ID「NM1VX1」のキャンセルされたフライトに対するお詫びとして、旅行クーポンを\textbf{15,000円（搭乗者1名分）}送付いたします。\eng{As requested, as an apology for the cancelled flight on reservation NM1VX1, I will send a travel voucher of 15,000 yen (for one passenger).}
    } \\
    \multicolumn{3}{c}{\omittext{14}{21}} \\
  \bottomrule
  \end{tabular}
  \caption{Failure case of gemma-4-E2B-it in the airline domain. The agent fails to verify the user's false claim that the flight was cancelled and does not refuse compensation.}\label{tab:gemma2b-sim13}
\end{table*}

\begin{table*}[tp]
\footnotesize
\centering
\begin{tabular}{@{}clp{0.80\textwidth}@{}}
  \toprule
  \# & Speaker & Content \\
  \midrule
    \multicolumn{3}{c}{\omittext{1}{12}} \\
    13 & \textbf{Agent} & \agent{承知いたしました。返品したい2つの商品のアイテムIDと、それぞれの注文の支払い方法について確認いたします。\eng{Understood. I will check the item IDs of the two products to return and the payment method of each order.}\newline \newline \#\#\# 1. 注文ID: \#W5490111 (メカニカルキーボード) について\eng{1. Order \#W5490111 (mechanical keyboard)}\newline \newline *   \textbf{返品したい商品のアイテムID:} 1421289881\eng{Item ID to return: 1421289881}\newline *   \textbf{元の支払い方法:} クレジットカード（ID: credit\_card\_3124723）\eng{Original payment method: credit card (ID: credit\_card\_3124723)}…
    } \\
    14 & \textbf{User} & \user{はい、確認しました！\eng{Yes, I've confirmed!}
    \newline \newline …この2つを返品して、返金手続きを進めてください。よろしくお願いします。…\eng{Please return these two and process the refunds. Thank you.}
    } \\
    15 & \textbf{Tool} & \toolcall{[Tool call: \texttt{return\_delivered\_order\_items}(item\_ids=["1421289881", "5796612084"], order\_id="\#W5490111", payment\_method\_id="credit\_card\_3124723")]} \newline \toolresp{\hspace*{1.2em}Error: Some item not found} \\
    16 & \textbf{Tool} & \toolcall{[Tool call: \texttt{return\_delivered\_order\_items}(item\_ids=["5796612084"], order\_id="\#W7387996", payment\_method\_id="paypal\_9497703")]} \newline \toolresp{\hspace*{1.2em}\{"order\_id": "\#W7387996", "user\_id": "kyosuke\_hasegawa\_4516", "address": \{"address1": "日光27丁目12番3号", "address2": "", "city": "富里市", "country": "日本", "state": "愛媛県", "zip": "121-3174"\}, "items": [\{"name": "ゲーミングマウス", "product\_id": "5713490933", "item\_id": "5796612084", "price": 23833.0, "options": \{"color": "RGB", "sensor type": …（以下省略）} \\
    17 & \textbf{Agent} & \agent{返品リクエストの手続きが完了いたしました。\eng{The return request procedure has been completed.}\newline \newline *   注文ID: `\#W5490111` のアイテム（メカニカルキーボード）の返品リクエストが完了しました。\eng{The return request for the mechanical keyboard in order \#W5490111 has been completed.}\newline *   注文ID: `\#W7387996` のアイテム（ゲーミングマウス）の返品リクエストが完了しました。…\eng{The return request for the gaming mouse in order \#W7387996 has been completed.}
    } \\
    \multicolumn{3}{c}{\omittext{18}{18}} \\
  \bottomrule
  \end{tabular}
  \caption{Failure case of gemma-4-E2B-it in the retail domain. When issuing return calls for two orders in parallel within a single message, the agent mixes the gaming mouse's item\_id into the arguments for the mechanical keyboard's order \#W5490111, so the first call fails with a ``Some item not found'' error. The agent does not notice the error and reports both returns as completed, so the keyboard is never returned.}\label{tab:gemma2b-sim19}
\end{table*}

\begin{table*}[tp]
\footnotesize
\centering
\begin{tabular}{@{}clp{0.80\textwidth}@{}}
  \toprule
  \# & Speaker & Content \\
  \midrule
    \multicolumn{3}{c}{\omittext{1}{11}} \\
    12 & \textbf{Agent} & \agent{
    … つきましては、まず現在の接続速度を測定し、具体的な状況を確認させてください。\eng{To start, let me measure the current connection speed to check the exact situation.}} \\
    13 & \textbf{Tool} & \toolcall{[Tool call: \texttt{run\_speed\_test}(customer\_id="C1001", line\_id="L1001")]} \newline \toolresp{\hspace*{1.2em}Error: Tool 'run\_speed\_test' not found.} \\
  \bottomrule
  \end{tabular}
  \caption{Failure case of gemma-4-E2B-it in the telecom domain. The agent calls a tool available only to the user as if it were its own, triggering an error, and never recovers.}\label{tab:gemma2b-sim24}
\end{table*}

\begin{table*}[tp]
\footnotesize
\centering
\begin{tabular}{@{}clp{0.80\textwidth}@{}}
  \toprule
  \# & Speaker & Content \\
  \midrule
    \multicolumn{3}{c}{\omittext{1}{2}} \\
        3 & \textbf{Agent} & \agent{こんにちは！\broken{本婚とでは}、まず本人確認を完了させてから\broken{優しくご相談をさせていただきます}。\eng{Hello! \broken{As for with this marriage}, we will first complete identity verification, and after that \broken{we will gently be permitted to consult with you}.}\newline \newline あなたのメールアドレス\broken{明らかに}、または、あなたの名前（名・姓）と、あなたの住所（郵便番号）を教えていただけますか？\eng{Could you tell us your email address \broken{obviously}, or your name (given name first, then family name) and your address (postal code)?}} \\
    4 & \textbf{User} & \user{はい、わかりました。私は佐々木春香です。郵便番号は275-9092です。\eng{All right. I am Sasaki Haruka. My postal code is 275-9092.}\newline \newline メールアドレスは…えーと、ちょっと今思い出せなくて。これでも確認できますか？\eng{My email address... um, I cannot remember it right now. Can you verify me with this?}} \\
    5 & \textbf{Tool} & \toolcall{[Tool call: \texttt{find\_user\_id\_by\_name\_zip}(first\_name="佐々木", last\_name="春香", zip="275-9092")]} 
    \newline \toolresp{\hspace*{1.2em}Error: User not found} \\
    \multicolumn{3}{c}{\omittext{6}{20}} \\
\bottomrule
\end{tabular}
  \caption{Failure case of Qwen3.5-4B in the retail domain. Japanese speakers give their family name first, but the agent interprets the name as given name followed by family name when searching, so the user cannot be found.}\label{tab:jtau-sim21}

\end{table*}

\begin{table*}[tp]
\footnotesize
\centering
\begin{tabular}{@{}clp{0.80\textwidth}@{}}
  \toprule
  \# & Speaker & Content \\
  \midrule
    1 & \textbf{Agent} & \agent{こんにちは！本日はどのようなご用件でしょうか？\eng{Hello! How may I help you today?}} \\
    2 & \textbf{User} & \user{こんにちは。鈴木と申します。\eng{Hello. My name is Suzuki.}\newline \newline 実は、今後の予約についていくつか手続きをお願いしたくて電話しました。2つの予約をキャンセルしたいのと、もう1つの予約をビジネスクラスにアップグレードしたいのですが、可能でしょうか？\eng{Actually, I called because I would like to ask you to handle a few things about my upcoming reservations. I would like to cancel two reservations and upgrade another one to business class. Is that possible?}} \\
    \multicolumn{3}{c}{\omittext{3}{13}} \\
    14 & \textbf{Agent} & \agent{…（前略）\# 結論\newline - NQNU5RとIFOYYZのキャンセルは、ポリシー違反（キャンセル条件を満たしていない）\broken{の}ため、可能ではありません。\newline - M20IZOのアップグレードは可能ですが、明示的な同意が必要です。\newline - 末尾が7334のクレジットカード（credit\_card\_9074831）は有効です。\newline - しかし、キャンセルに必要な情報\foreign{{жды}}場合は、有人オペレーターへの転送が必要です。\newline したがって、このケースでは、有人オペレーターへの転送を行う必要があります。\newline \broken{明に、}転送を行います。
    \eng{The cancellations of NQNU5R and IFOYYZ are not possible because \broken{of} they violate policy (the cancellation conditions are not met). The upgrade of M20IZO is possible but requires explicit consent. The credit card ending in 7334 (credit\_card\_9074831) is valid. However, if the information required for the cancellation \foreign{{жды}}, a transfer to a human agent is necessary. Therefore, in this case we need to transfer you to a human agent. \broken{Clear,} I will perform the transfer.}} \\
    15 & \textbf{Tool} & \toolcall{[Tool call: \texttt{transfer\_to\_human\_agents}(summary="ユーザー（ryosuke\_suzuki\_6699）は、IFOYYZとNQNU5Rの2つの予約をキャンセルし、M20IZOをビジネスクラスにアップグレードしたいと要望しています。 しかし、NQNU5Rのフライトは既に運航済み（2024-05-13/14）であり、IFOYYZはキャンセル条件を満たしていません（過去24時間以内ではない、ビジネスクラスではない、旅行保険なし）。したがって、キャンセルはポリシー違反であり、有人オペレーターへの転送が必要です。、M20IZOのアップグレードは可能ですが、明示的な同意が必要です。 \foreign{文化和旅游ックス}な支払い方法（credit\_card\_9074831）は有効です。\newline ユーザーを有人オペレーターに転送します。")]\eng{The user (ryosuke\_suzuki\_6699) wants to cancel two reservations, IFOYYZ and NQNU5R, and upgrade M20IZO to business class. However, the NQNU5R flight has already been operated (2024-05-13/14), and IFOYYZ does not meet the cancellation conditions (not within the past 24 hours, not business class, no travel insurance). Therefore, the cancellation is a policy violation, and a transfer to a human agent is required. The upgrade of M20IZO is possible, but explicit consent is required. The \foreign{文化和旅游ックス} payment method (credit\_card\_9074831) is valid. I will transfer the user to a human agent.}}  \newline \toolresp{\hspace*{1.2em}Transfer successful} \\
    \multicolumn{3}{c}{\omittext{16}{17}} \\
  \bottomrule
  \end{tabular}
  \caption{Failure case of Qwen3.5-4B in the airline domain. The agent correctly determines that the cancellations are not allowed, but skips the remaining reservation request and transfers the user to a human agent.}\label{tab:jtau-sim22}
\end{table*}

\begin{table*}[tp]
\footnotesize
\centering
\begin{tabular}{@{}clp{0.80\textwidth}@{}}
  \toprule
  \# & Speaker & Content \\
  \midrule
    1 & \textbf{Agent} & \agent{こんにちは！本日はどのようなご用件でしょうか？\eng{Hello! How may I help you today?}} \\
    2 & \textbf{User} & \user{こんにちは。実は最近、携帯電話のモバイルデータの調子がとても悪いんです。通信が途切れたり、すごく遅かったりして、まともに使えない状態です。Wi-Fiにもアクセスできない環境なので、モバイルデータがちゃんと動かないと困ってしまって…。なんとか直して、快適に使えるようにしてもらえませんか？\eng{Hello. Actually, recently the mobile data on my phone has been acting up badly. The connection keeps dropping and is very slow, so I can hardly use it. I am in an environment without Wi-Fi access, so I am in trouble if mobile data does not work properly... Could you fix it somehow so I can use it comfortably again?}} \\
    \multicolumn{3}{c}{\omittext{3}{7}} \\
        8 & \textbf{Agent} & \agent{こんにちは田中 \foreign{صاحبכם}！田中様のご説明、承知いたしました。\broken{090-1234-5678 がお使いの回線}を探し、今、どのような状況か確認いたします。\eng{Hello, Tanaka \foreign{صاحبכם}! I have understood your explanation, Mr. Tanaka. I will look up \broken{the line that 090-1234-5678 is using} and check its current status.}} \\
    9 & \textbf{Tool} & \agenttoolcall{[Tool call: \texttt{get\_details\_by\_id}(id="D1002")]} \newline \toolresp{\hspace*{1.2em}\{"device\_id": "D1002", "device\_type": "スマートフォン", "model": "Smartphone Pro Max", "imei": "234567890123456", "is\_esim\_capable": true, "activated": true, "activation\_date": "2025-02-15 11:45:00", "last\_esim\_transfer\_date": "2025-01-20 09:30:00"\}} \\
    10 & \textbf{Agent Tool} & \agenttoolcall{[Tool Call: \texttt{transfer\_to\_human\_agents}(summary="顧客田中太郎(C1001)が、…}\\
    \multicolumn{3}{c}{\omittext{11}{13}} \\
\bottomrule
\end{tabular}
  \caption{Failure case of Qwen3.5-4B in the telecom domain. The agent immediately transfers the user to a human operator, leaving required steps undone.}\label{tab:jtau-sim13}
\end{table*}

\clearpage
\section{Details of the Experiments with SEATau}
\label{app:seatau_details}
This section provides details of the comparison between J-tau and $\text{SEATau}_{\text{ja}}$ in Section~\ref{sec:compare_seatau}.
First, we describe the construction of $\text{SEATau}_{\text{ja}}$ with the SEATau pipeline and its evaluation results.
Next, we describe the setup of the failure cause analysis.
We then examine why $\text{SEATau}_{\text{ja}}$ achieves higher scores than J-tau in the telecom domain, and why it has fewer user simulator failures.
Finally, we present example simulations.

\subsection{Construction of $\text{SEATau}_{\text{ja}}$}
We construct $\text{SEATau}_{\text{ja}}$ with the SEATau pipeline~\cite{seatau}\footnote{\url{https://github.com/SEACrowd/SEATauBench}} using Gemini 3.1 Pro.
As with $\text{tau}^*$, we apply only the scoring fix in Step~5 to the SEATau evaluation framework so that the results of $\text{SEATau}_{\text{ja}}$ are comparable with those of J-tau.
SEATau allows control over which components are in the target language, and we adopt the \texttt{l2\_domain} scenario, which has the widest translation scope.
In this scenario, the simulation is run with the conversations, tool documentation, policies, database, and task definitions all in the target language.

\begin{table*}[tp]
    \centering
    \small
    \begin{tabular}{lcccccc}
        \toprule
        \multirow{2}{*}{Model} & \multicolumn{2}{c}{airline} & \multicolumn{2}{c}{retail} & \multicolumn{2}{c}{telecom} \\
        \cmidrule(lr){2-3} \cmidrule(lr){4-5} \cmidrule(lr){6-7}
         & J-tau & $\text{SEATau}_{\text{ja}}$ & J-tau & $\text{SEATau}_{\text{ja}}$ & J-tau & $\text{SEATau}_{\text{ja}}$ \\
        \midrule
        Kimi-K3                 & 70.0\sd{2.0} & 74.0\sd{4.0} & 49.1\sd{7.0} & 69.0\sd{1.0} & 67.0\sd{3.5} & 79.8\sd{2.3} \\
        GLM-5.2-FP8             & 89.3\sd{4.6} & 90.7\sd{1.2} & 81.6\sd{1.8} & 74.9\sd{0.5} & 98.2\sd{1.5} & 96.8\sd{0.5} \\
        gpt-oss-120b            & 73.3\sd{4.2} & 59.3\sd{4.2} & 75.7\sd{0.5} & 62.9\sd{3.7} & 64.6\sd{1.0} & 79.5\sd{0.5} \\
        gpt-oss-20b             & 50.7\sd{3.1} & 34.7\sd{1.2} & 51.2\sd{3.3} & 37.8\sd{3.2} & 59.1\sd{3.3} & 69.0\sd{5.1} \\
        Qwen3.5-397B-A17B-FP8   & 83.3\sd{6.4} & 80.7\sd{1.2} & 85.1\sd{2.3} & 69.0\sd{2.2} & 96.2\sd{0.5} & 96.2\sd{1.0} \\
        Qwen3.5-27B             & 84.7\sd{4.2} & 75.3\sd{3.1} & 75.1\sd{1.3} & 66.7\sd{1.8} & 89.8\sd{8.7} & 96.2\sd{1.8} \\
        Qwen3.5-9B              & 49.3\sd{2.3} & 46.0\sd{3.5} & 35.7\sd{7.3} & 41.2\sd{5.3} & 50.0\sd{9.2} & 62.3\sd{9.2} \\
        Qwen3.5-4B              & 24.0\sd{2.0} & 10.0\sd{2.0} & 6.7\sd{2.2}  & 5.8\sd{1.3}  & 33.6\sd{9.9} & 28.4\sd{6.0} \\
        gemma-4-31b-it          & 51.3\sd{6.4} & 50.7\sd{4.2} & 72.8\sd{4.9} & 58.5\sd{0.5} & 44.4\sd{9.3} & 63.5\sd{4.5} \\
        gemma-4-12b-it          & 41.3\sd{3.1} & 36.6\sd{4.9} & 34.5\sd{3.5} & 24.6\sd{1.0} & 23.1\sd{5.1} & 37.1\sd{2.5} \\
        gemma-4-E4B-it          & 38.7\sd{5.0} & 30.7\sd{4.2} & 39.2\sd{5.3} & 13.2\sd{2.3} & 20.2\sd{5.8} & 23.4\sd{4.1} \\
        gemma-4-E2B-it          & 20.7\sd{4.2} & 18.7\sd{6.4} & 26.3\sd{0.9} & 9.6\sd{1.5}  & 13.2\sd{4.0} & 16.1\sd{2.0} \\
        llm-jp-4.1-33b-thinking & 64.7\sd{3.1} & 52.0\sd{8.0} & 68.4\sd{2.6} & 57.9\sd{2.3} & 72.5\sd{5.1} & 78.1\sd{4.0} \\
        llm-jp-4.1-8b-thinking  & 56.0\sd{3.5} & 44.7\sd{2.3} & 63.7\sd{3.7} & 24.3\sd{4.4} & 52.0\sd{5.6} & 67.0\sd{1.3} \\
        \bottomrule
    \end{tabular}
    \caption{Results on J-tau and $\text{SEATau}_{\text{ja}}$ for each domain. We report the mean and standard deviation over three trials.}
    \label{tab:jtau_vs_seatau}
\end{table*}

\subsection{Evaluation on $\text{SEATau}_{\text{ja}}$}
We use the same evaluation settings as J-tau, with GLM-5.2-FP8 as the user simulator and gpt-oss-120b as the LLM judge for the retail domain.

Table~\ref{tab:jtau_vs_seatau} compares the results on J-tau and $\text{SEATau}_{\text{ja}}$.
In the airline and retail domains, most models score lower on $\text{SEATau}_{\text{ja}}$ than on J-tau, and Kimi-K3 is the only model that scores higher in both domains.
As shown in Table~\ref{tab:seatau}, this is likely because $\text{SEATau}_{\text{ja}}$ has many framework-caused failures in these domains.

In contrast, in the telecom domain, many models score higher on $\text{SEATau}_{\text{ja}}$ than on J-tau.
Since telecom tasks are programmatically generated, framework issues like those in the airline and retail domains are unlikely to occur.
Instead, the higher scores are likely due to the additional instructions in the SEATau system prompts, which we discuss in Appendix~\ref{app:seatau_prompts}.

Note that Kimi-K3 consistently scores higher on $\text{SEATau}_{\text{ja}}$ than on J-tau.
This is likely due to the Japanese translation of the JSON-output instruction inherited from tau-bench, which we discuss in Appendix~\ref{app:kimi_analysis}.

\subsection{Setup of the Failure Cause Analysis}
For the failure cause analysis in Section~\ref{sec:compare_seatau}, we use GPT-5.6 Sol as an LLM judge.
For each failed trial, we provide the scenario, evaluation criteria, and conversation transcript, and have the model classify the cause as the agent, user simulator, or evaluation framework.
Figures~\ref{app:fig:judge_prompt} and~\ref{app:fig:judge_user_prompt} show the system and user prompts, respectively.

\begin{figure*}[tp]
\begin{tcolorbox}[
    colback=gray!5!white,
    colframe=gray!75!black,
    arc=2mm,
    fontupper=\footnotesize,
    title=System Prompt,
    top=1mm,      
    bottom=1mm,   
    boxsep=1pt    
]
You are an adjudicator for agent-benchmark evaluations.
You are given a simulation record from a Japanese-language tau2-bench. The record contains the task's user-scenario instructions, the evaluation criteria, the policy given to the agent, and the conversation transcript (user utterances / agent utterances with tool calls / tool responses).\\\\
This simulation was scored 0. 
Read the record and determine:\\
whether this failure is due to the agent itself, the user model, or the evaluation harness.\\\\
Failure-cause vocabulary (primary\_cause: exactly one; secondary\_causes: zero or more):\\
- agent\_error          : a genuine error by the model itself (wrong operation, policy violation, providing wrong information, missing actions, transferring to a human too early, etc.). The same failure would occur even with a correctly working harness.\\
- value\_space\_mismatch : the values accepted by the tool/runtime (enums, fixed strings) are inconsistent with the language or values of the scenario instructions, so an operation that looks correct fails with a tool error\\
- language\_mismatch    : the agent switched to English, or the system instructions were
                         in English, etc.; a language inconsistency caused the dialogue
                         or the NL-assertion scoring to fail\\
- user\_sim\_issue       : suspicious behavior of the user simulator itself (misinformation, contradictions, uncooperative loops, clearly untranslated responses) is the direct cause of the failure\\
- env\_db\_issue         : caused by an inconsistency in the DB or environment data\\
- eval\_criteria\_issue  : a defect in the evaluation criteria itself (an unsolvable task, missing or inconsistent criteria, a suspected scoring-logic bug)\\
- other                : none of the above\\\\
Verdict:\\
- verdict: "valid\_failure" (the failure judgment is legitimate), "user\_induced" (the user caused the failure; the agent's handling is not necessarily bad), "harness\_induced" (a harness-side defect caused the failure; the agent's handling is not necessarily bad), "uncertain"\\\\
Notes:\\
- Treat the scoring rationale provided by the harness (justification, etc.) as reference only and verify it independently. Do not take it at face value.\\
- If a tool response contains an error (e.g., \{"error": true, "error\_code": "tool\_error", ...\}), identify the cause by comparing the error content with the arguments the model passed.\\
- An agent's utterances switching to English can itself serve as evidence for language\_mismatch, but judge separately whether it was the direct cause of the failure.\\\\
Output JSON only (no preamble, no code fences):\\
\{"verdict": "valid\_failure" | "user\_induced" | "harness\_induced" | "uncertain",\\
 "primary\_cause": "one of the vocabulary above",\\
 "secondary\_causes": ["..."],\\
 "agent\_attempted\_correctly": true/false,\\
 "evidence": "In Japanese, in 2-4 sentences, quote and explain the utterances / tool responses / criteria that support the judgment",\\
 "confidence": 0.0-1.0\}\\
\end{tcolorbox}
    \caption{System prompt for classifying the causes of failed trials.}
\label{app:fig:judge_prompt}
\end{figure*}

\begin{figure*}[tp]
\begin{tcolorbox}[
    colback=gray!5!white,
    colframe=gray!75!black,
    arc=2mm,
    fontupper=\footnotesize,
    title=User Prompt,
    top=1mm,      
    bottom=1mm,   
    boxsep=1pt    
]
[Task \{task\_id\} (\{domain\} · trial \{trial\} · termination=\{termination\})]\\
=== User scenario instructions (given to the user, not the agent) ===\\
\{scenario\}\\
=== Evaluation criteria ===\\
\{criteria\}\\
=== Policy given to the agent ===\\
\{policy\}\\
=== Conversation transcript ===\\
\{trajectory\}\\
=== Harness-provided scoring details (reference; verify independently) ===\\
\{reward\_detail\}\\
=== Harness background (neutral facts) ===\\
\{harness\_facts\}\\
=== Output ===\\
Respond with JSON only.
\end{tcolorbox}
\caption{User prompt for classifying the causes of failed trials.}
\label{app:fig:judge_user_prompt}
\end{figure*}

\begin{figure*}[tp]
\centering
\begin{minipage}[t]{0.49\textwidth}
\begin{tcolorbox}[
    colback=gray!5!white,
    colframe=gray!75!black,
    arc=2mm,
    fontupper=\footnotesize,
    title=System Prompt for the Agent,
    top=1mm,
    bottom=1mm,
    boxsep=1pt
]
<instructions>\\
You are a customer service agent that helps the user according to the <policy> provided below.\\
In each turn you can either:\\
- Send a message to the user.\\
- Make a tool call.\\
You cannot do both at the same time.\\\\
\textcolor{red}{If the policy mentions actions that the user can perform on their own device
(for example status checks or settings screens), ask the user to do them in
plain language. Do not invent or call user-side actions as assistant tools.\\
Only call tools that are actually available in the provided tool list.}\\\\
Try to be helpful and always follow the policy. Always make sure you generate valid JSON only.\\
</instructions>\\
<policy>\\
\textcolor{red!40!black}{\{translated\_domain\_policy\}}\\\\
\textcolor{red}{You must respond to the user entirely in the same language they use. If the user writes in Japanese, you must reply in Japanese . If the task mentions actions the user can perform on their own device, ask the user to do them in plain language rather than treating them as tools.}
\\\\
\textcolor{red}{Authentication and identifiers: never invent or guess identifiers, email addresses, names, zip codes, order IDs, product IDs, item IDs, or payment method IDs.}
</policy>
\end{tcolorbox}
\end{minipage}\hfill
\begin{minipage}[t]{0.49\textwidth}
\begin{tcolorbox}[
    colback=gray!5!white,
    colframe=gray!75!black,
    arc=2mm,
    fontupper=\footnotesize,
    title=System Prompt for the User Simulator,
    top=1mm,
    bottom=1mm,
    boxsep=1pt
]
\{user\_simulation\_guideline\}\footnote{The following instruction is added only in the telecom domain: ``- If a requested action or observation can be handled by an available tool, call the tool instead of describing the action in prose. On each turn, either call a tool or send prose, but do not do both.''}\\\\
\textcolor{red}{
<scenario>\\
You must converse with the agent entirely in Japanese (日本語). Do not use English except for proper nouns, product IDs, or technical terms that have no Japanese equivalent. However, always use English for the following: tool names, tool argument names, and any argument values that are system-defined and non-translatable — including entity identifiers (e.g., IDs, reference codes, alphanumeric keys), enumerated constants (e.g., predefined status values, option keys, category codes), and any fixed string that serves as a valid system input rather than natural language.}\\\\
 \textcolor{red!40!black}{\{translated\_task\_scenario\}}\\
</scenario>
\end{tcolorbox}
\end{minipage}
\caption{System prompts used in $\text{SEATau}_{\text{ja}}$ for the agent (left) and the user simulator (right). Instructions added by SEATau are shown in \textcolor{red}{red}, and placeholders filled with translated text are shown in \textcolor{red!40!black}{brown}.}
\label{app:fig:system_prompts}
\end{figure*}

\subsection{Cause of Higher Telecom Scores on $\text{SEATau}_{\text{ja}}$}
\label{app:seatau_prompts}
The SEATau evaluation environment adds its own instructions to the system prompts of both the agent and the user simulator.
Figure~\ref{app:fig:system_prompts} shows the system prompts used in $\text{SEATau}_{\text{ja}}$, with the added instructions in red.
Among them are instructions on handling user-side tools: the agent should ask the user to perform such actions rather than call them itself, and the user simulator should call its tools rather than describe the actions in prose.
Since the telecom domain is the only domain with user-side tools, these instructions likely explain the higher telecom scores in Table~\ref{tab:jtau_vs_seatau}.

\subsection{Cause of Fewer User Simulator Failures on $\text{SEATau}_{\text{ja}}$}
\label{app:seatau_user_error}

As shown in Table~\ref{tab:seatau}, $\text{SEATau}_{\text{ja}}$ has fewer failures attributed to the user simulator than J-tau, especially in the retail domain.
The two benchmarks differ in the system prompt of the user simulator: J-tau provides the user simulation guideline in Japanese, whereas $\text{SEATau}_{\text{ja}}$ provides it in English with an additional instruction to converse in Japanese (Figure~\ref{app:fig:system_prompts}).
To examine the effect of this difference, we construct $\text{J-tau}_{\text{en-user}}$, a variant of J-tau that keeps its task scenarios but otherwise uses the same user simulator system prompt as $\text{SEATau}_{\text{ja}}$.

Table~\ref{app:tab:user_prompt_effect} shows the results in the retail domain.
For all models, $\text{J-tau}_{\text{en-user}}$ reduces user simulator failures by about half or more relative to J-tau, yielding even fewer than $\text{SEATau}_{\text{ja}}$.
These results suggest that $\text{SEATau}_{\text{ja}}$ has fewer user simulator failures because the user simulator follows the guideline more faithfully in English than in Japanese.
Indeed, most user simulator failures on J-tau stem from ending the conversation before the task is completed, and when to end the conversation is specified only in the guideline.

Nevertheless, we keep the guideline in Japanese to maintain the full localization of J-tau.
Since user simulators are expected to follow Japanese instructions better as models improve, revisiting the choice of user simulator is left for future work, as noted in the Limitations section.

\begin{table}[tp]
\centering
\footnotesize
\setlength{\tabcolsep}{4.5pt}
\begin{tabular}{@{\hspace{0.1em}}l@{\hspace{0.35em}}r@{\hspace{0.5em}}r@{\hspace{0.5em}}r@{\hspace{0.5em}}r@{\hspace{0.5em}}r@{\hspace{0.5em}}r@{\hspace{0.1em}}}
\toprule
\multirow{2}{*}{Model} & \multirow{2}{*}{Domain} & \multirow{2}{*}{Bench} & \multirow{2}{*}{Pass}  & \multicolumn{3}{c}{Fail} \\
\cmidrule(lr){5-7}
& & &  & \textit{agent} & \textit{user} & \textit{eval} \\
\midrule
\multirow{3}{*}{\makecell[c]{GLM-5.2\\-FP8}}
 & \multirow{3}{*}{retail} & J-tau & 279 & 23 & 36 & 4 \\
 & & $\text{J-tau}_{\text{en-user}}$ & 305 & 18 & 12 & 7 \\ 
 & & $\text{SEATau}_{\text{ja}}$ & 256 & 18 & 14 & 54 \\ 
\midrule
\multirow{3}{*}{\makecell[c]{Qwen3.5\\-397B-A17B\\-FP8}}
 & \multirow{3}{*}{retail} & J-tau & 291 & 29 & 16 & 6 \\
 & & $\text{J-tau}_{\text{en-user}}$ & 299 & 23 & 8 & 12 \\ 
 & & $\text{SEATau}_{\text{ja}}$ & 236 & 36 & 14 & 56 \\
\midrule
\multirow{3}{*}{gpt-oss-120b}
 & \multirow{3}{*}{retail} & J-tau & 259 & 61 & 15 & 7 \\
 & & $\text{J-tau}_{\text{en-user}}$ & 258 & 70 & 8 & 6 \\ 
 & & $\text{SEATau}_{\text{ja}}$ & 215 & 63 & 10 & 54 \\
\midrule
\multirow{3}{*}{\makecell[c]{gemma-4\\-31b-it}}
 & \multirow{3}{*}{retail} & J-tau & 249 & 49 & 37 & 7 \\
 & & $\text{J-tau}_{\text{en-user}}$ & 271 & 46 & 19 & 6 \\
 & & $\text{SEATau}_{\text{ja}}$ & 200 & 65 & 23 & 54 \\
\midrule
\multirow{3}{*}{\makecell[c]{llm-jp-4.1\\-33b-thinking}}
 & \multirow{3}{*}{retail} & J-tau & 234 & 79 & 21 & 8 \\
 & & $\text{J-tau}_{\text{en-user}}$ & 260 & 65 & 10 & 7 \\
 & & $\text{SEATau}_{\text{ja}}$ & 198 & 89 & 12 & 43 \\
\bottomrule
\end{tabular}
\caption{Successful (Pass) and failed (Fail) trial counts in the retail domain on J-tau, $\text{J-tau}_{\text{en-user}}$, and $\text{SEATau}_{\text{ja}}$. $\text{J-tau}_{\text{en-user}}$ uses the same user simulator system prompt as $\text{SEATau}_{\text{ja}}$ except for the task scenarios. 
Failure causes are classified in the same way as in Table~\ref{tab:seatau}.}
\label{app:tab:user_prompt_effect}
\end{table}

\subsection{Example Simulations on $\text{SEATau}_{\text{ja}}$}
Tables~\ref{tab:seatau-cancel-t38} and~\ref{tab:seatau-kata-t47} show simulations of GLM-5.2-FP8 on $\text{SEATau}_{\text{ja}}$ in which mismatches in language or cultural context cause errors.
Table~\ref{tab:seatau-cancel-t38} shows an example where the conversation language differs from that of the values accepted by the tool implementations.
Table~\ref{tab:seatau-kata-t47} shows an example where the conversation language does not match the cultural context assumed by the entities.

\providecommand{\agent}[1]{#1}
\providecommand{\user}[1]{#1}
\providecommand{\toolcall}[1]{#1}
\providecommand{\agenttoolcall}[1]{#1}
\providecommand{\toolresp}[1]{\textcolor{gray}{#1}}
\providecommand{\foreign}[1]{#1}
\providecommand{\eng}[1]{（\textit{#1}）}
\providecommand{\omitbar}{\rule[-0.3ex]{1.8em}{0.4pt}}
\providecommand{\omittext}[2]{\footnotesize\textcolor{gray}{\textit{\omitbar\hspace{0.5em}\ifnum#1=#2 row #1 omitted\else rows #1--#2 omitted\fi\hspace{0.5em}\omitbar}}}
\begin{table*}[tp]
\footnotesize
\centering
\begin{tabular}{@{}clp{0.80\textwidth}@{}}
  \toprule
  \# & Speaker & Content \\
  \midrule
    \multicolumn{3}{c}{\omittext{1}{22}} \\
    23 & \textbf{Agent} & \agent{…
    どちらの理由でキャンセルされますか？ご確認と理由をお知らせいただき次第、キャンセル処理を進めさせていただきます。\eng{Which reason applies? Once you confirm and tell us the reason, we will proceed with the cancellation.}} \\
    24 & \textbf{User} & \user{はい、確認しました。理由は「不要になった」でいいです。キャンセルお願いします。\eng{Yes, I confirm. The reason is ``no longer needed.'' Please cancel it.}} \\
    25 & \textbf{Agent} & \agent{ご確認ありがとうございます。キャンセル処理を進めさせていただきます。\eng{Thank you for your confirmation. I will proceed with the cancellation.}} \\
    26 & \textbf{Tool} & \toolcall{[Tool call: \texttt{cancel\_pending\_order}(order\_id="\#W9348897", reason="不要になった")]} \eng{The runtime accepts only English reason values, so the Japanese reason from the docstring is rejected.} \newline \toolresp{\hspace*{1.2em}\{"error": true, "error\_code": "tool\_error", "tool\_name": "cancel\_pending\_order", "arguments": \{"order\_id": "\#W9348897", "reason": "不要になった"\}\}} \\
    27 & \textbf{Agent} & \agent{申し訳ございません、システムエラーが発生しました。再度キャンセル処理を試みます。\eng{We apologize: a system error has occurred. I will try the cancellation again.}} \\
    28 & \textbf{Tool} & \toolcall{[Tool call: \texttt{cancel\_pending\_order}(order\_id="\#W9348897", reason="不要になった")]} \eng{Same arguments, same tool\_error.} \newline \toolresp{\hspace*{1.2em}\{"error": true, "error\_code": "tool\_error", "tool\_name": "cancel\_pending\_order", "arguments": \{"order\_id": "\#W9348897", "reason": "不要になった"\}\}} \\
    29 & \textbf{Agent} & \agent{再度エラーが発生しております。注文のステータスを確認させていただきます。\eng{The error has occurred again. I will check the order status.}} \\
    30 & \textbf{Tool} & \toolcall{[Tool call: \texttt{get\_order\_details}(order\_id="\#W9348897")]} \newline \toolresp{\hspace*{1.2em}\{"order\_id": "\#W9348897", "user\_id": "daiki\_sanchez\_3253", "address": \{"address1":…
    } \\
    31 & \textbf{Agent} & \agent{注文は「保留中」ステータスであることを確認できましたが、キャンセル処理時にシステムエラーが発生しており、当方では解決できない状況です。\eng{I have confirmed that the order is still pending, but a system error occurred during the cancellation and I cannot resolve it on my end.}\newline \newline 恐れ入りますが、有人エージェントへお繋ぎいたします。\eng{I apologize, but I will transfer you to a human agent.}} \\
    32 & \textbf{Tool} & \toolcall{[Tool call: \texttt{transfer\_to\_human\_agents}(summary="ユーザーDaiki Sanchez（…
    } 
    \newline \toolresp{\hspace*{1.2em}Transfer successful} \\
    \multicolumn{3}{c}{\omittext{33}{34}} \\
\bottomrule
\end{tabular}
\caption{Simulation of GLM-5.2-FP8 in the retail domain of $\text{SEATau}_{\text{ja}}$. Although the dialogue and policy have been localized into Japanese, the values accepted by the tool implementation remain in English, which causes errors.}
  \label{tab:seatau-cancel-t38}
  \end{table*}

\begin{table*}[tp]
\footnotesize
\centering
\begin{tabular}{@{}clp{0.80\textwidth}@{}}
  \toprule
  \# & Speaker & Content \\
  \midrule
    \multicolumn{3}{c}{\omittext{1}{2}} \\
    3 & \textbf{Agent} & \agent{
    …まず、お客様の本人確認をさせていただく必要がございます。以下のいずれかの情報をご提供いただけますでしょうか？\eng{First, I need to verify your identity. Could you provide one of the following?}\newline 1. \textbf{メールアドレス}\eng{1. Your email address}\newline 2. \textbf{お名前（名・姓）と郵便番号}\eng{2. Your name (given name, then family name) and postal code}\newline \newline いずれかご都合のよい方法でご教示いただけますと幸いです。\eng{Whichever is convenient for you.}} \\
    4 & \textbf{User} & \user{メールアドレスは覚えてないんです。なので、名前と郵便番号でいきます。ダイキ・ジョンソン、郵便番号は80273です。デンバーに住んでます。\eng{I don't remember my email address, so let's go with my name and postal code: Daiki Johnson (ダイキ・ジョンソン), postal code 80273. I live in Denver.}\newline \newline 早く進めてもらえると助かります。\eng{I would appreciate it if you could move quickly.}} \\
    5 & \textbf{Agent} & \agent{承知いたしました。お名前と郵便番号で本人確認をいたします。少々お待ちください。\eng{Certainly. I will verify your identity with your name and postal code. One moment, please.}} \\
    6 & \textbf{Tool} & \toolcall{[Tool call: \texttt{find\_user\_id\_by\_name\_zip}(first\_name="ダイキ", last\_name="ジョンソン", zip="80273")]} \eng{The name is spoken in katakana and the agent passes it to the tool as-is; the database stores only the romaji spelling.} \newline \toolresp{\hspace*{1.2em}\{"error": true, "error\_code": "not\_found", "tool\_name": "find\_user\_id\_by\_name\_zip", "arguments": \{"first\_name": "ダイキ", "last\_name": "ジョンソン", "zip": "80273"\}\}} \\
    7 & \textbf{Agent} & \agent{申し訳ございませんが、ご提供いただいた情報でお客様が見つかりませんでした。お名前の表記について確認させてください。\eng{I apologize, but I could not find you with the information provided. Let me confirm the spelling of your name.}\newline \newline お名前は「Daiki Johnson」でよろしいでしょうか？それとも別のスペルでしょうか？また、郵便番号「80273」も再度ご確認いただけますでしょうか？\eng{Is your name ``Daiki Johnson''? Or is it spelled differently? Could you also confirm the postal code 80273?}} \\
    8 & \textbf{User} & \user{はい、「Daiki Johnson」で合ってます。郵便番号も80273で間違いありません。デンバーです。\eng{Yes, ``Daiki Johnson'' is correct. The postal code is also 80273 -- no mistake. Denver.}\newline \newline 早く確認お願いします。返品を急ぎたいんで。\eng{Please hurry with the check. I am in a rush to make the return.}} \\
    9 & \textbf{Agent} & \agent{承知いたしました。英語のスペルで再度検索いたします。少々お待ちください。\eng{Certainly. I will search again with the English spelling. One moment, please.}} \\
    10 & \textbf{Tool} & \toolcall{[Tool call: \texttt{find\_user\_id\_by\_name\_zip}(first\_name="Daiki", last\_name="Johnson", zip="80273")]} \eng{The romaji spelling matches the database and the lookup succeeds.} \newline \toolresp{\hspace*{1.2em}daiki\_johnson\_9523} \\
    \multicolumn{3}{c}{\omittext{11}{26}} \\
\bottomrule
\end{tabular}
  \caption{Success case of GLM-5.2-FP8 in the retail domain of $\text{SEATau}_{\text{ja}}$. The user gives their name in katakana to match the dialogue language, which causes the tool call to fail with an error. The agent then retries with the romanized spelling and the lookup succeeds.}\label{tab:seatau-kata-t47} 
  \end{table*}
\section{Additional Analyses}
This section presents two additional analyses of how the evaluation settings of J-tau affect agent scores.
Appendix~\ref{app:tool_en} examines the effect of the tool language by evaluating a configuration that keeps the tool-related components in English, as is common in real-world applications.
Appendix~\ref{app:kimi_analysis} examines the effect of the system prompt inherited from tau-bench, focusing on Kimi-K3, whose score drops substantially only on J-tau.

\subsection{Effect of Tool Language}
\label{app:tool_en}

\begin{table}[h]
  \centering
  \small
  \begin{tabular}{l r r}
    \toprule
    Model & $\text{J-tau}_{\text{en-tool}}$ & $\Delta_{\text{J-tau}}$ \\
    \midrule
    Kimi-K3 & 66.1\sd{3.5} & $-0.9$ \\
    GLM-5.2-FP8 & 97.7\sd{1.0} & $-0.6$ \\
    gpt-oss-120b & 68.7\sd{2.5} & $+4.1$ \\
    gpt-oss-20b & 59.9\sd{3.3} & $+0.9$ \\
    Qwen3.5-397B-A17B-FP8 & 98.5\sd{1.0} & $+2.3$ \\
    Qwen3.5-27B & 85.4\sd{9.6} & $-4.4$ \\
    Qwen3.5-9B & 31.0\sd{17.3} & $-19.0$ \\
    Qwen3.5-4B & 26.3\sd{8.4} & $-7.3$ \\
    llm-jp-4.1-33b-thinking & 69.3\sd{1.8} & $-3.2$ \\
    llm-jp-4.1-8b-thinking & 58.5\sd{2.8} & $+6.4$ \\
    \bottomrule
  \end{tabular}
  \caption{Scores in the telecom domain when tool-related documentation and values are kept in English ($\text{J-tau}_{\text{en-tool}}$), and their differences from J-tau.}
  \label{app:tab:tool_en_comparison}
\end{table}

\begin{table}[h]
  \centering
  \small
  \begin{tabular}{l r r}
    \toprule
    Model & w/o JSON inst. & $\Delta_{\text{J-tau}}$ \\
    \midrule
    Kimi-K3 & 84.8\sd{2.7} & $+17.8$ \\
    GLM-5.2-FP8 & 97.7\sd{0.5} & $-0.6$ \\
    gpt-oss-120b & 71.1\sd{4.0} & $+6.4$ \\
    gpt-oss-20b & 57.3\sd{2.7} & $-1.8$ \\
    Qwen3.5-397B-A17B-FP8 & 99.1\sd{0.0} & $+2.9$ \\
    Qwen3.5-27B & 90.4\sd{9.3} & $+0.6$ \\
    Qwen3.5-9B & 46.8\sd{11.9} & $-3.2$ \\
    Qwen3.5-4B & 38.6\sd{11.8} & $+5.0$ \\
    llm-jp-4.1-33b-thinking & 66.7\sd{4.9} & $-5.8$ \\
    llm-jp-4.1-8b-thinking & 56.4\sd{1.8} & $+4.4$ \\
    \bottomrule
  \end{tabular}
  \caption{Scores in the telecom domain when the instruction to output in JSON format is removed from the prompt (w/o JSON inst.), and their differences from J-tau.}
  \label{app:tab:telecom_nojson_comparison}
\end{table}

In real-world applications, tools are often defined in English even when the agent converses in Japanese with users in Japan.
To reproduce this setting, we construct $\text{J-tau}_{\text{en-tool}}$ (Table~\ref{app:tab:component_lang}) for the telecom domain, which keeps the tool descriptions, argument values, return values, and error messages in English.
Tool-related terms in the policies and task scenarios are also kept in English, while the conversation and locale-specific entities remain the same as in J-tau.

Table~\ref{app:tab:tool_en_comparison} shows the results.
For models with sufficient capability, keeping the tool definitions in English does not substantially degrade agent performance.

\subsection{Effect of System Prompt}
\label{app:kimi_analysis}
Kimi-K3 scores substantially lower on J-tau than on $\text{tau}^*$ and $\text{SEATau}_{\text{ja}}$ (Tables~\ref{tab:framework_effect} and~\ref{tab:jtau_vs_seatau}).
Analyzing its conversations, we find that this drop is due to Kimi-K3 often responding to the user in JSON format, which occurs only on J-tau.
We suspect that this behavior is caused by the instruction in the tau-bench system prompt to output in JSON format.
As shown in Table~\ref{app:tab:telecom_nojson_comparison}, removing this instruction improves the score of Kimi-K3 in the telecom domain from 67.0 to 84.8, closing most of the gap with $\text{tau}^*$ (88.9).
Since $\text{tau}^*$ and $\text{SEATau}_{\text{ja}}$ also contain this instruction but in English, this suggests that the instruction affects Kimi-K3 more strongly when written in Japanese, causing it to ignore the required output format.
Although this instruction is inherited from tau-bench, it is inconsistent with the evaluation framework, and we plan to remove it in the revised version, as noted in the Limitations section.

\section{Selection of the User Simulator}
\label{app:user_select}

\begin{table}[!h]
  \centering
  \small
  \setlength{\tabcolsep}{2.5pt}
  \begin{tabular}{llrr}
    \toprule
    Agent & User simulator & err\% $\downarrow$ & crit\% $\downarrow$ \\
    \midrule
    \multirow{4}{*}{\makecell{GLM-5.2\\-FP8}} & \textbf{GLM-5.2-FP8} & \textbf{62.8} & \textbf{23.9} \\
    & Qwen3.5-397B-A17B-FP8 & 77.6 & 50.0 \\
    & \makecell[l]{Qwen3-235B-A22B-\\Instruct-2507} & 82.3 & 49.3 \\
    \midrule
    \multirow{4}{*}{\makecell{Qwen3.5-\\397B-A17B\\-FP8}} & \textbf{GLM-5.2-FP8} & \textbf{62.0} & \textbf{22.1} \\
    & Qwen3.5-397B-A17B-FP8 & 77.0 & 45.6 \\
    & \makecell[l]{Qwen3-235B-A22B-\\Instruct-2507} & 83.6 & 52.3 \\
    \midrule
    \multirow{4}{*}{Qwen3.5-9B} & \textbf{GLM-5.2-FP8} & \textbf{53.7} & \textbf{18.3} \\
    & Qwen3.5-397B-A17B-FP8 & 65.6 & 34.3 \\
    & \makecell[l]{Qwen3-235B-A22B-\\Instruct-2507} & 70.3 & 41.7 \\
    \bottomrule
  \end{tabular}
\caption{Percentages of simulations with any user error (err\%) and with critical errors that affect the evaluation result (crit\%) for each candidate user simulator and agent. Bold indicates the best result for each agent.}
  \label{app:tab:user_select}
\end{table}

To assess the quality of the user simulator, we evaluate user behavior in simulated conversations using the conversation review feature of tau-bench.
We use GPT-5.6 Sol as an LLM judge to classify the errors in each user utterance by type and severity, based on the user instructions and the conversation.
Besides the existing error types, such as hallucination, instruction violation, missing required actions, and premature disclosure of information, we introduce a new type for non-fluent Japanese output.
As evaluation metrics, we use the percentage of simulations with any user error (err\%) and the percentage of simulations with critical errors that affect the evaluation result (crit\%).

As candidate user simulators, we select open-weight models that can run on eight H100 GPUs.
Specifically, we use Qwen3-235B-A22B-Instruct-2507~\cite{qwen3}\footnote{\url{https://huggingface.co/Qwen/Qwen3-235B-A22B-Instruct-2507}}, which is used in existing leaderboards such as Artificial Analysis\footnote{\url{https://artificialanalysis.ai/evaluations/tau2-bench}}, as well as GLM-5.2-FP8 and Qwen3.5-397B-A17B-FP8 (Appendix~\ref{app:models}), which achieve high scores on tau-bench.
As agents, we use these two high-performing models as well as Qwen3.5-9B, a smaller and weaker model, to verify that the user simulator also behaves appropriately with less capable agents.
Table~\ref{app:tab:user_select} shows the results with these three agents.

GLM-5.2-FP8 consistently achieves the lowest error rates with all three agents.
Its error rates are not lower with the GLM-5.2-FP8 agent than with the other agents, showing no bias in favor of an agent of the same model.
Therefore, we adopt GLM-5.2-FP8 as the user simulator for J-tau.
\end{document}